\documentclass[10pt,twocolumn,letterpaper]{article}

\usepackage[pagenumbers]{cvpr} 

\usepackage{multirow}

\usepackage[table]{xcolor}

\newcolumntype{D}{>{\columncolor{blue!4}}c}
\newcolumntype{H}{>{\columncolor{blue!8}}c}
\usepackage{capt-of}
\newcommand{\DynVal}[1]{\multicolumn{1}{D}{#1}}
\newcommand{\HighVal}[1]{\multicolumn{1}{H}{#1}}

\usepackage{amsfonts,bm,makecell,algorithm,algpseudocode}
\usepackage{cuted} 
\usepackage{iftex}
\ifPDFTeX\else
\usepackage{fontspec}
\fi

\definecolor{cvprblue}{rgb}{0.21,0.49,0.74}
\usepackage[pagebackref,breaklinks,colorlinks,allcolors=cvprblue]{hyperref}

\def\confName{CVPR}
\def\confYear{2027}

\title{Dusting the Phantom: Gaussian Allocation for Fading Dynamic Content}

\author{
Minhyeok Roh$^{1}$,
Minseo Lee$^{1}$,
Jun Young Jeong$^{2}$,
Byeonghyeon Lee$^{1}$, \\
Lucas Yunkyu Lee$^{3}$,
Jaesik Park$^{4}$,
Eunbyung Park$^{1\dagger}$ \\[4pt]
$^{1}$Yonsei University \quad
$^{2}$Electronics and Telecommunications Research Institute (ETRI) \\
$^{3}$Pohang University of Science and Technology (POSTECH) \quad
$^{4}$Seoul National University
}

\begin{document}
\maketitle
\begin{strip}
    \centering
    \includegraphics[width=1\linewidth]{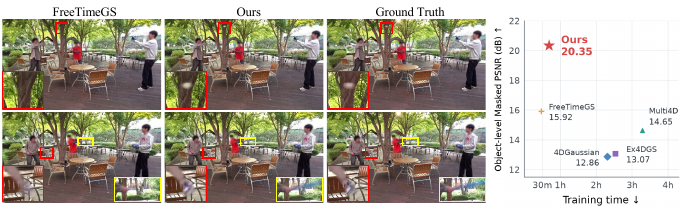}
    \captionsetup{hypcap=false}
    \captionof{figure}{\textbf{Reconstruction under highly challenging dynamics.}
    \textbf{Top}: The rapidly moving ball is missing from the reconstructions of a strong baseline model, FreeTimeGS, whereas our method recovers it. \textbf{Bottom}: Under severe motion blur, the bat and glove are barely visible in competing reconstructions but are more clearly recovered by our method. \textbf{Right}: Object-level Masked PSNR across four objects on the Custom dataset, evaluated within manually annotated object regions under severe motion blur (\cref{object-level analysis} for details). Our method achieves the highest object-level PSNR with a training time comparable to FreeTimeGS.}
    \label{fig:teaser}
\captionsetup{hypcap=true}
\end{strip}

\begin{abstract}
Dynamic Gaussian representations enable high-quality, real-time volumetric video, with FreeTimeGS standing out as a high-performing approach. Nevertheless, challenging motion remains a persistent issue in dynamic Gaussian reconstruction: small fast-moving objects may disappear, and fine textures in highly dynamic regions are often lost. In particular, simply training longer or using more Gaussians may still fail to recover small, fast-moving objects. We argue that these failures stem not from limited representational capacity alone, but also from where and when Gaussian primitives are allocated. To address these limitations, we augment FreeTimeGS with two Gaussian allocation strategies. First, Seed Injection identifies under-reconstructed dynamic regions using temporal appearance changes and reconstruction errors, then reinitializes low-opacity Gaussians at geometrically plausible 3D locations inferred from multi-view observations, independently of their previous locations. Second, Temporal Biased Relocation uses time-resolved optimization signals to guide the spatiotemporal placement of relocated Gaussians. Both strategies maintain a fixed Gaussian budget without modifying the underlying FreeTimeGS representation. Experiments across multiple dynamic-scene datasets demonstrate improved reconstruction quality, particularly for small, rapidly moving objects and fine appearance details in highly dynamic regions.~{\urlstyle{same}Project page: \url{https://minhyeok2001.github.io/Dusting-the-Phantom/}}
\end{abstract}

\section{Introduction}

\begin{figure}[t]
    \centering
    \includegraphics[
        width=\linewidth,
        trim=0 50 0 0,
        clip
    ]{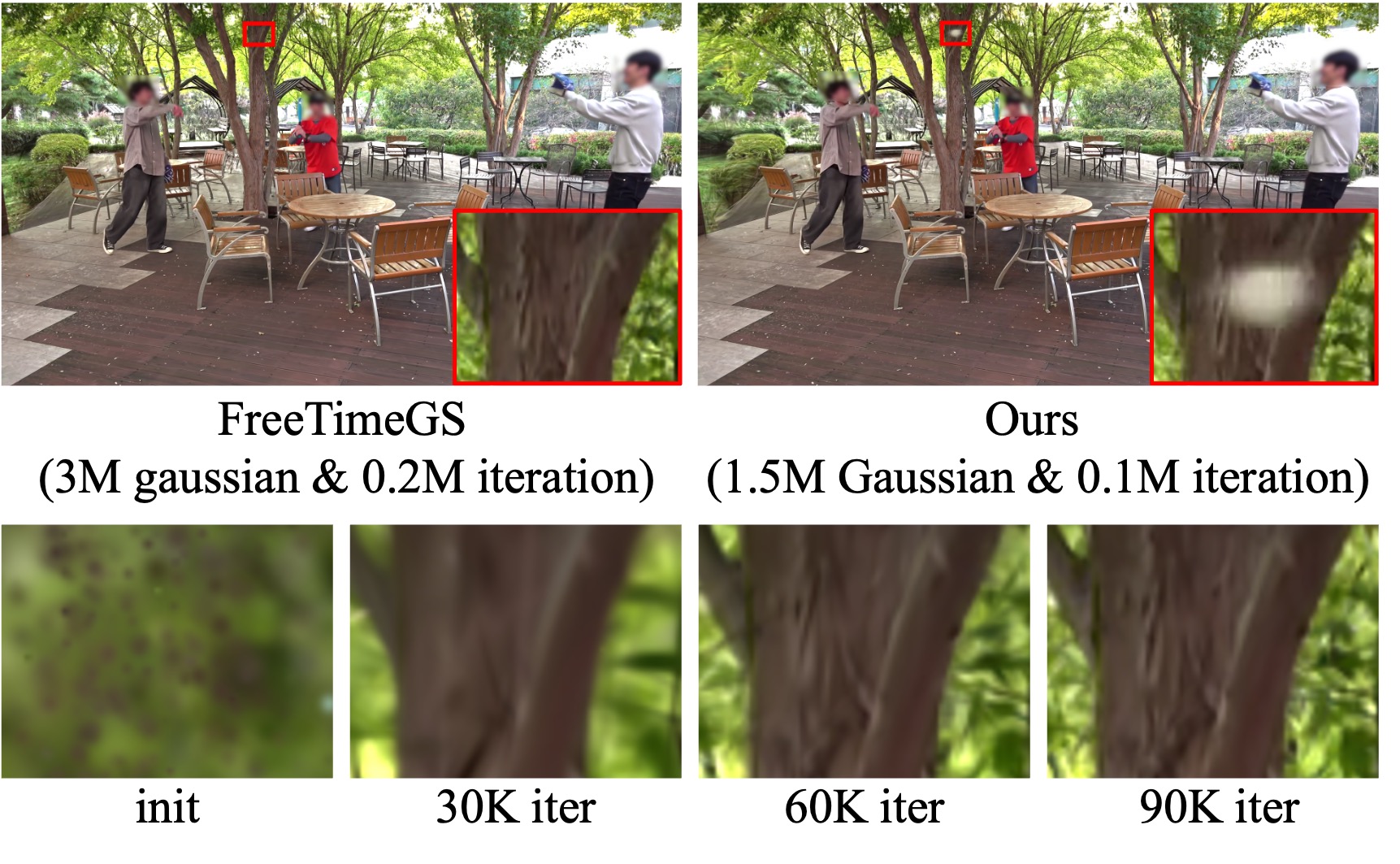}
    \caption{\textbf{Limitations of increasing Gaussian capacity and training iterations.}
    \textbf{Top}: FreeTimeGS fails to recover the fast-moving ball with 3M Gaussians and 0.2M iterations (left), whereas ours succeeds with 1.5M Gaussians and 0.1M iterations (right).
    \textbf{Bottom}: FreeTimeGS close-ups of the expected ball region at initialization and 30K, 60K, and 90K iterations. The ball remains missing throughout training.}
    \label{fig:intro_fig}
\end{figure}

Recent advances in 4D volumetric video promise to transform how we experience visual content: rather than watching events from fixed, recorded viewpoints, viewers can freely explore dynamic scenes as they unfold in space and time.
From free-viewpoint sports replays~\cite{chen2019fast,dai2024real} and immersive performances~\cite{collet2015high} to lifelike telepresence~\cite{Lawrence2021}, such experiences require faithful reconstruction of both appearance and motion, together with interactive rendering. Recent advances in Neural Radiance Fields (NeRFs)~\cite{mildenhall2021nerf} and 3D Gaussian Splatting (3DGS)~\cite{kerbl20233d} have brought this vision closer to reality through high-quality novel-view synthesis, with successful extensions to dynamic scenes. In particular, 3DGS's high-quality, real-time rendering makes it attractive for next-generation immersive media, driving interest in Gaussian-based 4D scene modeling.

Many approaches have been developed to extend 3DGS to dynamic scenes, modeling temporal variations through deformation-based models~\cite{yang2024deformable,wu20244d,huang2024sc,xu2024grid4d,bae2024per} or explicit spatiotemporal Gaussian representations~\cite{yang2024real,li2024spacetime,wang2025freetimegs,liao2026sharptimegs,lee2024fully,wang2026retimegs,yeom2026trigs}. Among these, FreeTimeGS~\cite{wang2025freetimegs} achieves high-quality dynamic scene reconstruction with real-time rendering. However, it still struggles with challenging dynamic content: small, rapidly moving objects may disappear entirely (\cref{fig:teaser}(Top)), while fine textures in highly dynamic regions are often poorly recovered (\cref{fig:teaser}(Bottom)). One possible approach is to train longer or increase the representational capacity by introducing more Gaussian primitives. While these strategies can improve overall reconstruction quality and may also help reconstruct finer details in dynamic regions, they come at additional computational cost. Yet, even with more Gaussians and longer training, the small, rapidly moving ball remains missing, as shown in the top row of \cref{fig:intro_fig}. This suggests that increasing capacity and extending training alone may be insufficient, motivating us to revisit how Gaussian primitives are allocated during optimization.

The bottom row of~\cref{fig:intro_fig} shows crops of the region where the white ball should appear at different optimization iterations. FreeTimeGS~\cite{wang2025freetimegs} initializes few or no Gaussians in this region and fails to recover the ball throughout training. This initialization failure may stem from RoMa~\cite{edstedt2024roma}, the dense matcher used in FreeTimeGS's preprocessing pipeline, struggling to establish reliable correspondences in small, motion-blurred, or textureless regions. Even if a region is missed during initialization, FreeTimeGS’s relocation strategy would be expected to allocate Gaussians to that region and recover the missing content. However, relocation determines new primitive locations based on existing Gaussians, limiting its ability to populate regions where few or no Gaussians were initialized. To address this limitation, we introduce \emph{Seed Injection}, which directly allocates Gaussians to under-reconstructed dynamic regions using synchronized multi-view observations. Specifically, it identifies these regions through temporal intensity changes and reconstruction errors, and reinitializes low-opacity Gaussians at geometrically plausible 3D locations obtained from multi-view geometry. By placing Gaussians independently of their previous locations, \emph{Seed Injection} enables the recovery of regions that conventional relocation struggles to populate, while preserving the total Gaussian budget.

However, spatial allocation through Seed Injection alone is not sufficient for dynamic scene reconstruction, as optimization signals can vary significantly across timestamps. This motivates us to revisit the temporal placement of relocated Gaussians in FreeTimeGS~\cite{wang2025freetimegs}. Its relocation strategy selects target Gaussians using scores aggregated across timestamps and assigns their temporal centers to relocated primitives. Consequently, temporal placement follows the targets' existing temporal centers without explicitly accounting for timestamp-specific optimization signals. We therefore introduce \emph{Temporal Biased Relocation}, which uses time-resolved optimization signals to guide temporal placement toward timestamps with stronger observed signals. Together with \emph{Seed Injection}, this approach improves reconstruction quality in dynamic scenes without increasing the total Gaussian count or the number of training iterations.

    \emph{Seed Injection} and \emph{Temporal Biased Relocation} improve the spatial and temporal allocation of Gaussian primitives. We apply both strategies to FreeTimeGS~\cite{wang2025freetimegs}, our baseline, without modifying its underlying representation. Our contributions are as follows: (i) We introduce \emph{Seed Injection}, which allocates Gaussians to poorly reconstructed dynamic regions independently of existing Gaussian locations. (ii) We propose \emph{Temporal Biased Relocation}, which leverages temporal optimization signals to guide the spatiotemporal placement of relocated Gaussians. (iii) Experiments across multiple dynamic-scene datasets demonstrate improved reconstruction quality, particularly in highly dynamic regions and for small, rapidly moving objects.
\section{Related Work}

\paragraph{Dynamic Gaussian Splatting.}
Building upon 3D Gaussian Splatting (3DGS)~\cite{kerbl20233d}, dynamic methods extend Gaussian representations over time through tracking or deformation~\cite{luiten2024dynamic,yang2024deformable,wu20244d,lu20243d}. Other approaches explicitly parameterize Gaussian evolution using 4D primitives, temporal interpolation, or continuous trajectories~\cite{yang2024real,li2024spacetime,lee2024fully,liao2026sharptimegs,yeom2026trigs,wang2026retimegs,lee2026freetimegs++}, while hybrid methods combine deformable Gaussians with transient 4D primitives~\cite{oh2025hybrid,wang2026multi4d}. FreeTimeGS~\cite{wang2025freetimegs}, which serves as our baseline, introduces flexible temporal lifespans to model Gaussian appearance and motion over time. Related to our goal of reconstructing challenging dynamic content, SPIN-4DGS~\cite{kim2026implicit} addresses reconstruction under large inter-frame displacements by reducing cross-frame interference through attribute prediction from explicitly collected spatiotemporal positions. Our work instead targets Gaussian allocation as a source of dynamic reconstruction failures, without modifying the underlying dynamic Gaussian representation.

\paragraph{Gaussian Densification and Relocation.}
The original 3DGS adaptively clones or splits Gaussians based on positional gradients, while subsequent works improve density control through alternative densification criteria~\cite{rota2024revising,zhang2024pixel,ye2024absgs,zhou2026gradient}. Beyond densification, 3DGS-MCMC~\cite{kheradmand20243d} introduces relocation to redistribute underutilized Gaussian capacity. Dynamic methods additionally incorporate temporal properties such as lifespan, velocity, and visibility into density control~\cite{liao2026sharptimegs,sandu2026temporally,yeom2026trigs,wang2026retimegs}. While these methods use temporal cues to guide primitive selection, the spatiotemporal destination of Gaussian capacity receives less consideration. Our method instead introduces new support in under-represented dynamic regions and uses timestamp-specific optimization signals to determine the temporal destination of relocated capacity.

\paragraph{Gaussian Allocation for Under-Reconstructed Regions.}
Error-based densification~\cite{rota2024revising} directly targets under-reconstructed regions using reconstruction errors, but remains dependent on existing primitives as densification sources. LPM~\cite{yang2025improving} alleviates this dependency by localizing error regions in 3D through cross-view feature correspondence, but relies on an external matching model, which may be less reliable in regions with weak visual correspondence. SpaceTimeGS~\cite{li2024spacetime} enables guided sampling even in regions with sparse or missing Gaussian support, but still relies on the current Gaussian representation to provide a coarse depth prior for localizing new primitives. In contrast, our approach directly lifts observations from under-reconstructed dynamic regions into 3D using synchronized multi-view geometry, without relying on existing Gaussian support or an external correspondence model.

\begin{figure*}[t!]
    \centering
    \includegraphics[width=1\linewidth]{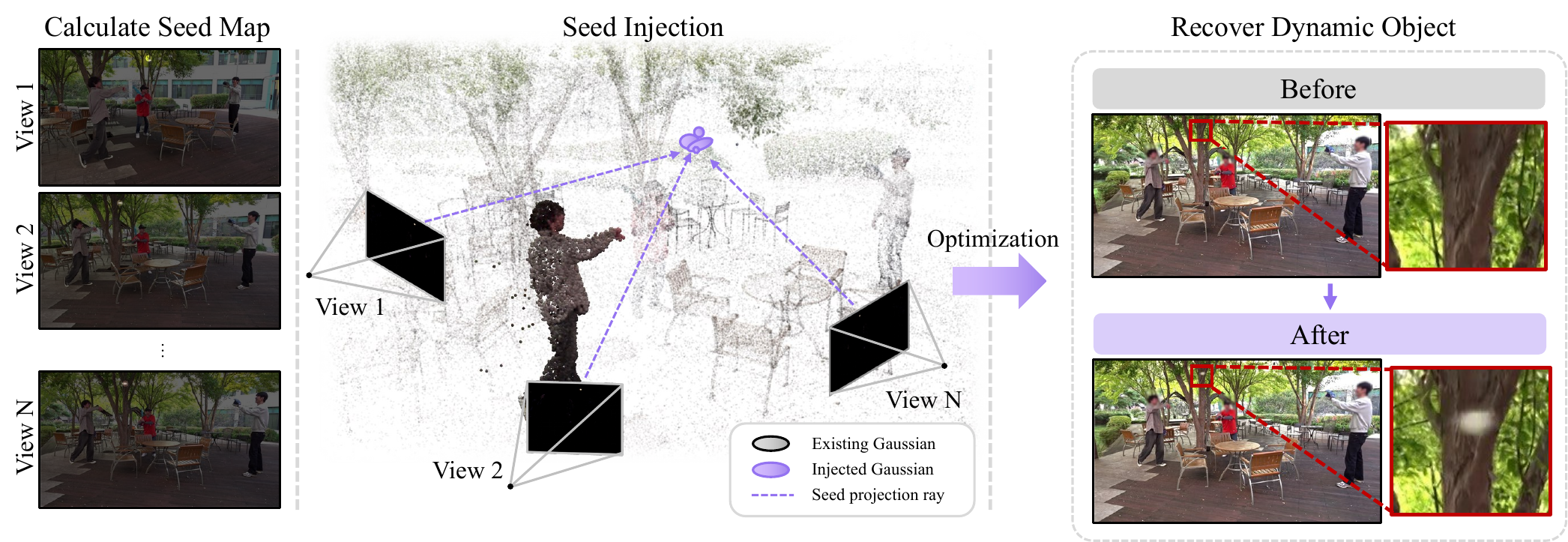}
    
    \caption{\textbf{Overview of Seed Injection.} We identify under-reconstructed dynamic regions from synchronized multi-view observations and sample candidate pixels from the resulting seed map. The sampled observations are lifted into 3D using multi-view geometry to reinitialize existing low-opacity Gaussians independently of their previous locations. After optimization, the injected Gaussians provide additional support for dynamic content that is poorly represented or missing in the original reconstruction.}
    \label{fig:seed_injection}
\end{figure*}

\section{Method}
\label{sec:method}

In this section, we present our approach for improving Gaussian allocation in dynamic scene reconstruction through \emph{Seed Injection} and \emph{Temporal Biased Relocation}. We first review the FreeTimeGS~\cite{wang2025freetimegs} representation and its Gaussian relocation strategy as preliminaries in ~\cref{sec:preliminaries}. Then, ~\cref{sec:seed_injection} describes how \emph{Seed Injection} reallocates existing Gaussians to under-reconstructed dynamic regions, and ~\cref{sec:temporal_relocation} details how \emph{Temporal Biased Relocation} redistributes Gaussian capacity across time based on temporal optimization signals. 

\subsection{Preliminaries}
\label{sec:preliminaries}
\paragraph{FreeTimeGS representation.}
FreeTimeGS~\citep{wang2025freetimegs} represents a dynamic scene using Gaussian primitives that can appear at arbitrary spatial locations and timestamps. Each Gaussian includes a spatial position $\mu_x$, temporal center $\mu_t$, duration $d$, velocity $v$, and opacity $\sigma$, along with standard 3DGS attributes. To model local motion, FreeTimeGS assigns each Gaussian a linear motion function,
\begin{equation}
\mu_x(t)
=
\mu_x + v(t-\mu_t),
\label{eq:freetime_motion}
\end{equation}
where $\mu_x(t)$ denotes the spatial position of the Gaussian at timestamp $t$. The contribution of each Gaussian over time is further controlled by a temporal opacity function,
\begin{equation}
\sigma(t)
=
\exp\left(
-\frac{1}{2}
\left(
\frac{t-\mu_t}{d}
\right)^2
\right).
\label{eq:freetime_temporal_opacity}
\end{equation}
The final opacity is obtained by multiplying the base opacity by $\sigma(t)$, where $\mu_t$ and $d$ control the temporal center and extent, respectively.

\paragraph{FreeTimeGS relocation.} FreeTimeGS maintains a fixed Gaussian budget by periodically recycling low-opacity Gaussians through relocation. Specifically, it defines a sampling score $s$ for each Gaussian primitive as
\begin{equation}
s
=
\lambda_g \nabla_g
+
\lambda_o \sigma,
\label{eq:ftgs_relocation_score}
\end{equation}
where $\nabla_g$ and $\sigma$ denote the spatial gradient and opacity of the Gaussian primitive, respectively, and $\lambda_g$ and $\lambda_o$ are their corresponding weights. At each relocation interval, low-opacity Gaussians are recycled by copying the parameters of existing Gaussians selected according to their sampling scores, while maintaining a fixed Gaussian budget.

\subsection{Seed Injection}
\label{sec:seed_injection}
As described in~\cref{sec:preliminaries}, FreeTimeGS~\cite{wang2025freetimegs} relocates Gaussians based on scores defined over existing primitives, making it difficult to populate poorly represented regions. To address this limitation, we aim to estimate new 3D Gaussian locations directly from multi-view image observations, independently of the existing Gaussian distribution. We therefore introduce \emph{Seed Injection}, which identifies candidate pixels in under-reconstructed dynamic regions using seed maps, and then estimates their 3D positions through multi-view triangulation based on viewing-ray proximity. The resulting new 3D points serve as reinitialization locations for existing low-opacity Gaussians. An overview of \emph{Seed Injection} is illustrated in ~\cref{fig:seed_injection}.

\paragraph{Dynamic Error-based Seed Map.}
To reallocate existing Gaussians to highly dynamic regions, we construct a seed map for each camera view $v$ at timestamp $t$, combining a reconstruction error map $E_{v,t}$ and a binary dynamic mask $D_{v,t}$. The reconstruction error map $E_{v,t}$ highlights poorly reconstructed regions but may also contain errors in static areas. To suppress these errors, we construct the dynamic mask $D_{v,t} \in \{0,1\}^{H\times W}$ from differences between adjacent frames, focusing on temporally varying regions. The per-pixel reconstruction error and resulting seed map $M_{v,t}$ are defined as
\begin{equation}
\begin{aligned}
E_{v,t}(p) &= \left\|I_{v,t}(p)-\hat{I}_{v,t}(p)\right\|_1,\\
M_{v,t}(p) &= D_{v,t}(p)E_{v,t}(p).
\end{aligned}
\label{eq:seed_error}
\end{equation}
Here, $I_{v,t}$ and $\hat{I}_{v,t}$ denote the ground-truth and rendered images, respectively, and $p$ denotes a pixel coordinate. The resulting seed map assigns high values to pixels that are both dynamic and poorly reconstructed. Further details are provided in the supplementary material.

\paragraph{Multi-view Seed Lifting.}
For each time step $t$, we select a subset $\mathcal{S} \subseteq \mathcal{V}$ of $C$ synchronized camera views, where $\mathcal{V}$ denotes the set of all available views. From each selected view $v \in \mathcal{S}$, we sample $N_{\mathrm{seed}}$ candidate seed pixels with probabilities proportional to their error values in the seed map $M_{v,t}$ and collect them as $\mathcal{P}_{v,t}=\{p_{v,t}^1,\ldots,p_{v,t}^{N_{\mathrm{seed}}}\}$, where $p_{v,t}^i \in \mathbb{R}^2$ denotes the 2D image coordinate of the $i$-th sampled seed candidate. Since each sampled seed provides only a 2D image location, its 3D location is ambiguous. To approximate the 3D location, we match seed candidates across views based on the proximity of their viewing rays, as candidates corresponding to the same or nearby 3D region are expected to produce rays that pass close to each other.

Each sampled seed pixel defines a viewing ray originating from its camera center. Let $o_v \in \mathbb{R}^3$ denote the camera center of view $v$, and $d_v^i \in \mathbb{R}^3$ the unit viewing direction corresponding to pixel $p_{v,t}^i$. The viewing ray is defined as
\begin{equation}
r_v^i(\lambda)=o_v+\lambda d_v^i,
\end{equation}
where $\lambda$ denotes the distance along the ray. For simplicity, we omit the time index $t$ in the ray notation. This yields $C \times N_{\mathrm{seed}}$ candidate rays, denoted by $\mathcal{R}_t = \{r_v^i \mid v \in \mathcal{S},\, i=1,\ldots,N_{\mathrm{seed}}\}$.

To identify candidates corresponding to nearby 3D regions, we compute the minimum distance between ray pairs from different views ($v \neq v'$). For each pair $(r_v^i,r_{v'}^j)$, we find the closest points by solving
\begin{equation}
\begin{aligned}
(\lambda_{ij}^{*},\gamma_{ij}^{*})
=
\arg\min_{\lambda,\gamma}
\left\|
r_v^i(\lambda)-r_{v'}^j(\gamma)
\right\|_2^2.
\end{aligned}
\label{eq:ray_closest_points}
\end{equation}
The resulting closest points are $q_v^{i,j}=r_v^i(\lambda_{ij}^{*})$ and $q_{v'}^{i,j}=r_{v'}^j(\gamma_{ij}^{*})$, with their distance given by $\delta_{ij}=\|q_v^{i,j}-q_{v'}^{i,j}\|_2$. For each ray, we select the valid cross-view partner with the smallest $\delta_{ij}$, treating the pair as potentially originating from the same 3D point.

Since noise may prevent rays corresponding to the same 3D point from intersecting exactly, we estimate the corresponding 3D seed position as the midpoint of their closest points,
\begin{equation}
x_{\mathrm{seed}}
=
\frac{1}{2}
\left(
q_v^{i,j}
+
q_{v'}^{i,j}
\right).
\end{equation}
At each timestep, the resulting 3D seed candidates are ranked in ascending order of their ray distances $\delta_{ij}$, and at most $N_{\max}$ seeds are retained for Gaussian injection. Further implementation details are provided in the supplementary material.

\paragraph{Gaussian Injection.}
At the midpoint of training, when reconstruction errors provide informative signals, we apply the seed lifting procedure described above to each frame in the training sequence using a single model snapshot, and cache the resulting 3D seeds. For each retained seed $x_{\mathrm{seed}}$, we reinitialize an available low-opacity Gaussian at the seed position. The number of reinitialized Gaussians at each timestep is limited by both the retained seeds and available low-opacity Gaussians, ensuring that the total number of Gaussians remains fixed throughout training. The scale of each reinitialized Gaussian is initialized isotropically based on the pixel footprint at the estimated depth, following prior works~\citep{barron2021mip,baranowski2026conegs}. Similar to FreeTimeGS~\citep{wang2025freetimegs}, we initialize its velocity from the displacement to the nearest lifted seed in the subsequent frame. This nearest-neighbor displacement serves only as a coarse motion initialization and does not assume temporal correspondence between seeds.

To stabilize the injected Gaussians, we temporarily disable relocation and process the training timestamps sequentially: we inject Gaussians at the first timestamp, perform 10 optimization iterations using only observations from that timestamp, then inject Gaussians at the second timestamp and perform another 10 iterations using only observations from the second timestamp, and so on. After completing the Gaussian Injection procedure, we resume the original frame-sampling schedule while keeping relocation disabled for an additional $10N_{\mathrm{frames}}$ iterations to allow the injected Gaussians to adapt to the scene, where $N_{\mathrm{frames}}$ denotes the number of training frames. Afterward, relocation is re-enabled according to the original schedule. All these iterations are included within the original training budget. Seed positions can be inaccurate due to noise, but half of the training budget remains after injection. During the remaining training, correctly placed Gaussians learn to represent fast-moving objects, while incorrectly placed Gaussians lose opacity and are later relocated. In our experiments, artifacts caused by inaccurate seed positions disappear during the remaining half of training. Further evidence of artifact disappearance is provided in the supplementary material.

\subsection{Temporal Biased Relocation}
\label{sec:temporal_relocation}

As described in ~\cref{sec:preliminaries}, FreeTimeGS~\citep{wang2025freetimegs} reuses Gaussians whose opacity falls below a predefined threshold and relocates them toward primitives with high sampling scores. However, FreeTimeGS pays limited attention to \textit{when} additional Gaussian capacity is needed, potentially leading to suboptimal temporal allocation. To address this limitation, we propose \emph{Temporal Biased Relocation}, which preserves timestamp-specific positional gradient magnitudes and uses them to determine the relocation time.

We follow FreeTimeGS~\citep{wang2025freetimegs} for selecting both the Gaussian to be relocated and the target Gaussian. However, once a target Gaussian $g$ is selected, its trajectory spans multiple timestamps, and we must determine \emph{when} along the trajectory the relocated Gaussian should be placed.
A large screen-space positional gradient at timestamp $t$ indicates that the reconstruction loss is sensitive to the Gaussian's projected position at that time. We use this sensitivity to guide the temporal placement of relocated Gaussians. We average the target Gaussian's screen-space positional gradient magnitudes over its observations at each timestamp during the current gradient accumulation window. Denoting this average magnitude by $\|\nabla_{g,t}\|_2$ and the set of timestamps at which $g$ was observed in this window by $\mathcal{T}_g$, we normalize these values to form a temporal relocation distribution:
\begin{equation}
P_g(t)
=
\frac{
\left\|\nabla_{g,t}\right\|_2
}{
\sum_{u\in\mathcal{T}_g}
\left\|\nabla_{g,u}\right\|_2
},
\qquad t\in\mathcal{T}_g.
\label{eq:conditional_time}
\end{equation}
Averaging before normalization prevents frequently sampled timestamps from dominating the relocation distribution solely due to repeated observations.
Then, we sample the relocation timestamp $t^* \sim P_g(t)$, favoring timestamps with larger average positional gradient magnitudes. We set the temporal center of the relocated Gaussian, $\hat{\mu}_t$, to the selected timestamp $t^*$, and its spatial center, $\hat{\mu}_x$, to the position of the target Gaussian $g$ at that time:
\begin{equation}
    \hat{\mu}_t = t^{*},
    \qquad
    \hat{\mu}_x = \mu_x(t^{*})
    = \mu_x + v(t^{*}-\mu_t).
    \label{eq:temporal_relocation_position}
\end{equation}
Here, $\mu_x$, $\mu_t$, and $v$ denote the spatial center, temporal center, and velocity of the target Gaussian, respectively.
We additionally adjust the temporal duration to account for the shifted temporal center, while the remaining parameters follow the original FreeTimeGS~\citep{wang2025freetimegs} relocation scheme.
Further details of the duration adjustment and relocation procedure are provided in the supplementary material.

\section{Experiments}
\label{sec:experiments}
\subsection{Experimental Setup}
\begin{table*}[t]
    \centering
    \footnotesize
    \caption{
        Quantitative comparison on the N3DV, CMU Panoptic,
        and Custom datasets.
        Best and second-best results are highlighted in
        \textbf{bold} and \underline{underlined}, respectively.
    }
    \label{tab:main_quantitative}
    \setlength{\tabcolsep}{2.6pt}
    \resizebox{\textwidth}{!}{%
    \begin{tabular}{@{}l*{3}{ccDH}@{}}
        \toprule
        & \multicolumn{4}{c}{N3DV}
        & \multicolumn{4}{c}{CMU Panoptic}
        & \multicolumn{4}{c}{Custom} \\
        \cmidrule(lr){2-5}
        \cmidrule(lr){6-9}
        \cmidrule(l){10-13}

        Method
        & PSNR$\uparrow$
        & LPIPS$\downarrow$
        & \makecell{Dynamic\\PSNR$\uparrow$}
        & \makecell{Highly Dynamic\\PSNR$\uparrow$}
        & PSNR$\uparrow$
        & LPIPS$\downarrow$
        & \makecell{Dynamic\\PSNR$\uparrow$}
        & \makecell{Highly Dynamic\\PSNR$\uparrow$}
        & PSNR$\uparrow$
        & LPIPS$\downarrow$
        & \makecell{Dynamic\\PSNR$\uparrow$}
        & \makecell{Highly Dynamic\\PSNR$\uparrow$} \\
        \midrule

        RealTime4DGS~\cite{yang2024real}
        & 31.92 & 0.091 & \underline{32.07} & \underline{22.92}
        & 27.87 & 0.110 & 25.32 & \underline{23.63}
        & 20.46 & 0.249 & 24.10 & 19.92 \\

        Ex4DGS~\cite{lee2024fully}
        & 30.75 & 0.087 & 29.87 & 22.10
        & 24.55 & 0.193 & 22.89 & 19.47
        & 19.58 & 0.306 & 22.06 & 17.45 \\

        4DGaussian~\cite{wu20244d}
        & 30.79 & 0.097 & 28.80 & 19.20
        & 26.18 & 0.191 & 22.10 & 19.59
        & 20.43 & 0.348 & 21.51 & 15.13 \\

        Multi4D~\cite{wang2026multi4d}
        & 31.46 & 0.096 & 30.47 & 21.25
        & 26.58 & 0.153 & 21.68 & 18.32
        & \underline{20.89} & 0.203 & 23.58 & 19.65 \\

        FreeTimeGS~\cite{wang2025freetimegs}
        & \underline{32.48} & \underline{0.042} & 31.44 & 22.32
        & \underline{28.55} & \underline{0.089} & \underline{26.24} & 23.33
        & 20.80 & \underline{0.169} & \underline{27.60} & \underline{24.07} \\


        \midrule

        \textbf{Ours}
        & \textbf{33.01} & \textbf{0.039} & \textbf{32.17} & \textbf{23.43}
        & \textbf{29.05} & \textbf{0.059} & \textbf{26.95} & \textbf{25.63}
        & \textbf{21.00} & \textbf{0.157} & \textbf{27.83} & \textbf{26.87} \\

        \bottomrule
    \end{tabular}%
    }
\end{table*}
\paragraph{Datasets.}    
We evaluate our method on three datasets with varying levels of scene dynamics. For N3DV~\cite{li2022neural}, we use all six scenes, each with 300 frames captured at 30 FPS by 19--21 cameras at $2704\times2028$ resolution. We downsample images by a factor of two, reserving camera 0 for testing and using the rest for training. For CMU Panoptic~\cite{joo2015panoptic}, we use six 150-frame sequences at 30 FPS and $640\times360$ resolution, with 27 training cameras and cameras 0, 10, 15, and 30 for testing. We set the test-view near plane to 0.5 to suppress artifacts from Gaussians near the test cameras. Finally, we introduce the Custom dataset, comprising Baseball and Break, to evaluate challenging dynamic scenes that involve complex motion alongside small, rapidly moving objects that are difficult to reconstruct and insufficiently represented in existing datasets. We capture both sequences using 23 synchronized cameras at $1920\times1080$ resolution and 30 FPS, with 250 and 300 frames, respectively, reserving one spatially intermediate camera for testing.

\paragraph{Evaluation.}
We evaluate reconstruction quality using PSNR, LPIPS~\cite{zhang2018unreasonable}, dynamic PSNR, and highly dynamic PSNR. The full results table, including DSSIM~\cite{wang2004image}, is provided in the supplementary material. For dynamic PSNR, we construct object-level masks in the test views using SAM3~\cite{carion2026sam} with dataset-specific prompts targeting moving objects. For the CMU Panoptic and Custom datasets in particular, we supplement these masks with manually annotated regions covering small, rapidly moving objects that off-the-shelf models~\cite{jocher2026ultralytics,karaev2025cotracker3,carion2026sam,zhang2026megaflow} fail to capture.
Because SAM3 segments entire objects without distinguishing their motion magnitude, these masks may include nearly static regions. For highly dynamic PSNR, we retain only pixels within the SAM3 masks whose optical-flow magnitudes exceed dataset-specific thresholds (5 pixels for CMU Panoptic and N3DV, and 10 pixels for Custom), using flow estimated by MegaFlow~\cite{zhang2026megaflow}, and then add the same manually annotated regions for CMU Panoptic and Custom datasets. Thus, dynamic PSNR evaluates the full segmented objects, whereas highly dynamic PSNR focuses on their high-motion regions; both include the manually annotated small, rapidly moving objects. Details of mask generation and filtering are provided in the supplementary material. All experiments use an NVIDIA RTX 5090 (32 GB), except RealTime4DGS~\cite{yang2024real}, which uses an NVIDIA RTX A6000 (48 GB) for both training and evaluation.

\paragraph{Implementation Details.}
We build on the reproduced FreeTimeGS~\cite{wang2025freetimegs} implementation and follow its default training configuration unless otherwise specified. We use 500K Gaussians for N3DV~\cite{li2022neural} and CMU Panoptic~\cite{joo2015panoptic}, and 1.5M Gaussians for the Custom dataset. Note that the total number of Gaussians in our model remains fixed even after \emph{Seed Injection}, which recycles existing Gaussians. We train for 30K iterations on N3DV, 50K on CMU Panoptic, and 100K on the Custom dataset, using a batch size of 1 throughout. The Custom dataset requires a longer training schedule due to its more challenging outdoor scenes and high-frequency background details. For \emph{Seed Injection}, we inject seeds once at the midpoint of training. We select $C=8$ synchronized cameras to construct the dynamic error-based seed maps, with $N_{\mathrm{seed}}=32$ candidates sampled per view and a maximum of $N_{\max}=256$ injected seeds per timestep. The velocity and scale are initialized as described in~\cref{sec:seed_injection}, while opacity and temporal duration are initialized to 0.05 and 0.03, respectively. \emph{Temporal Biased Relocation} is applied whenever relocation is performed throughout training.

\paragraph{Baseline Models.}
We compare our method with RealTime4DGS~\cite{yang2024real}, Ex4DGS~\cite{lee2024fully}, 4DGaussian~\cite{wu20244d}, Multi4D~\cite{wang2026multi4d}, and FreeTimeGS~\cite{wang2025freetimegs}. Since our method builds on FreeTimeGS, we match FreeTimeGS in all shared hyperparameters, including the number of Gaussians. SPIN-4DGS~\cite{kim2026implicit} also addresses challenging motion with large inter-frame displacements, but we exclude it from our comparisons because its official code is unavailable, making a controlled comparison difficult.

\begin{figure*}[t]
    \includegraphics[width=1\linewidth]{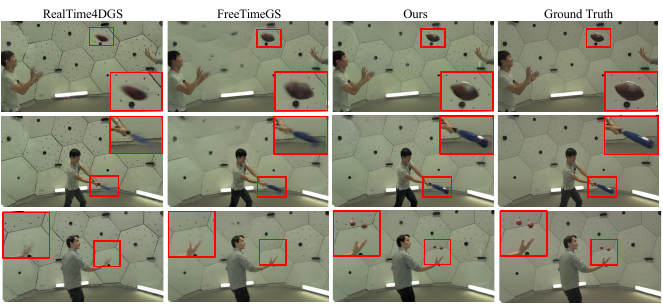}
    \caption{\textbf{Qualitative comparison on CMU Panoptic.} Our method better reconstructs moving balls and baseball bats, including juggling balls that are entirely missing from the reconstructions of baseline methods.}
    \label{fig:main_cmu}
\end{figure*}

\begin{figure*}[]
    \includegraphics[width=1\linewidth]{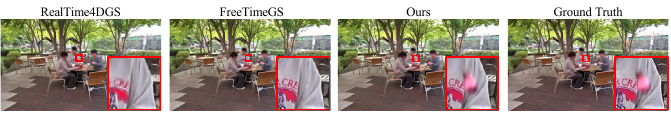}
    \caption{\textbf{Qualitative comparison on the Custom Break sequence.} Baseline methods fail to reconstruct the rapidly ejected pirate figure, whereas our method recovers it despite its small size and fast motion.}
    \label{fig:main_custom}
\end{figure*}

\subsection{Comparison Experiments}

\paragraph{N3DV.}
On N3DV~\cite{li2022neural}, which exhibits relatively mild scene dynamics, our method outperforms FreeTimeGS~\cite{wang2025freetimegs} in PSNR, dynamic PSNR, and highly dynamic PSNR (\cref{tab:main_quantitative}). The gains are more pronounced in dynamic and highly dynamic regions, suggesting that our allocation strategy remains effective even when overall scene motion is relatively mild. Qualitative comparisons are provided in the supplementary material.

\paragraph{CMU Panoptic.}
We evaluate our method on CMU Panoptic~\cite{joo2015panoptic}, which contains diverse and highly dynamic human motions. As shown in ~\cref{fig:main_cmu}, our method improves the reconstruction of moving balls and baseball bats, while recovering juggling balls that completely disappear in the RealTime4DGS~\cite{yang2024real} and FreeTimeGS~\cite{wang2025freetimegs} baselines. These improvements are also reflected in the quantitative results in ~\cref{tab:main_quantitative}.

\paragraph{Custom Dataset.}
We evaluate our method on the Custom dataset, consisting of Baseball and Break, with quantitative results reported in ~\cref{tab:main_quantitative}. In Baseball, FreeTimeGS~\cite{wang2025freetimegs} fails to reconstruct the rapidly moving ball and bat and loses fine details of the glove. Our method recovers the missing ball and better preserves the bat and glove details (\cref{fig:teaser}). In Break, the pirate figure rapidly ejected from a Pop-up Pirate toy is missing from the reconstructions of baseline methods, whereas our method successfully recovers it (\cref{fig:main_custom}).

\begin{figure*}[]
    \centering
    \includegraphics[width=1\linewidth]{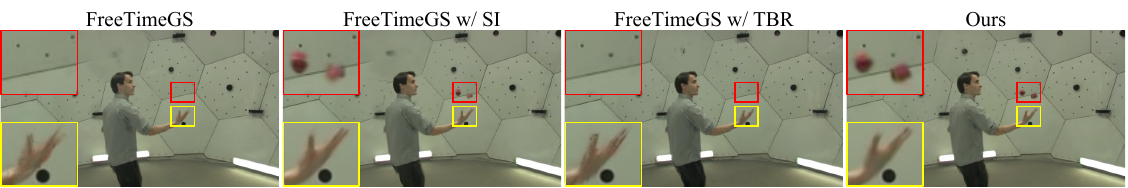}    
    \caption{\textbf{Qualitative ablation on the CMU Panoptic Juggle sequence.} SI and TBR denote \emph{Seed Injection} and \emph{Temporal Biased Relocation}, respectively. SI recovers missing balls, as highlighted in the red boxes, while TBR improves fine details such as the fingertips, as highlighted in the yellow boxes. Their combination provides the most complete reconstruction.}
    \label{fig:cmu_ablation_fig}
\end{figure*}

\begin{table}[h]
    \centering
    \caption{Ablation study of our proposed components on the CMU Juggle, Football, and Basketball sequences. SI and TBR denote \emph{Seed Injection} and \emph{Temporal Biased Relocation}, respectively.}
    \label{tab:cmu_ablation}
    \setlength{\tabcolsep}{3pt}
    \resizebox{\linewidth}{!}{%
    \begin{tabular}{lcDHc}
        \toprule
        Method
        & PSNR $\uparrow$
        & \makecell{Dynamic\\PSNR $\uparrow$}
        & \makecell{Highly Dynamic\\PSNR $\uparrow$}
        & \makecell{Training\\Time (m) $\downarrow$} \\
        \midrule
        FreeTimeGS~\cite{wang2025freetimegs}
        & 28.22 & 26.24 & 23.26 & \textbf{4.19} \\
        FreeTimeGS w/ SI
        & 28.53 & \underline{27.14} & \underline{25.49} & 4.29 \\
        FreeTimeGS w/ TBR
        & \underline{28.57} & 26.89 & 24.23 & \underline{4.26} \\
        \midrule
        \textbf{FreeTimeGS w/ SI\&TBR (Ours)}
        & \textbf{28.98} & \textbf{27.33} & \textbf{26.11} & 4.43 \\
        \bottomrule
    \end{tabular}%
    }
\end{table}

\subsection{Ablation Studies and Analysis}

\paragraph{Component Ablation.}
We conduct an ablation study on three highly dynamic CMU Panoptic~\cite{joo2015panoptic} sequences: Basketball, Football, and Juggle. All other training settings remain identical to those described above, except for the ablated components. As shown in \cref{fig:cmu_ablation_fig}, \emph{Seed Injection} recovers rapidly moving objects, such as the juggling balls, that the baseline fails to reconstruct, while \emph{Temporal Biased Relocation} improves fine details such as the fingertips, as well as under-reconstructed static regions. \cref{tab:cmu_ablation} reports the average results across the three sequences: Combining both components further improves reconstruction quality with only a 5.7\% increase in training time.
Further ablation studies on the key parameters of \emph{Seed Injection}, including $N_{\mathrm{seed}}$, $N_{\max}$, and the number of camera views, as well as the additional analysis of both \emph{Seed Injection} and \emph{Temporal Biased Relocation}, are provided in the supplementary material.

\begin{figure}[t]
    \centering
    \includegraphics[
        width=\columnwidth,
        trim={0.5cm 0 0 0},
        clip
    ]{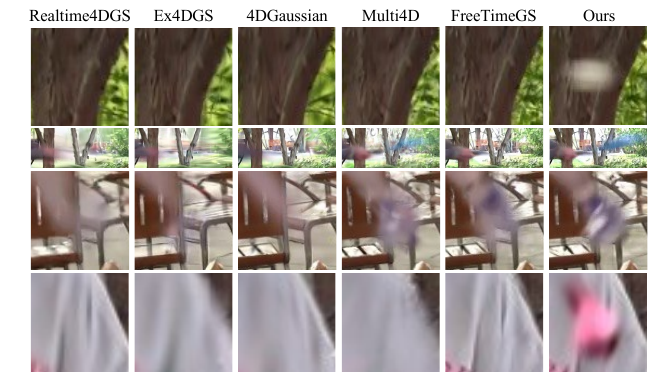}
    \caption{\textbf{Object-level qualitative comparison on the Custom dataset.} Our method recovers the ball, bat, and pirate figure that are barely visible or missing in competing reconstructions, while more clearly preserving the glove's shape under motion blur.}
    \label{fig:analysis_quali}
\end{figure}

\paragraph{Object-level Analysis on the Custom dataset.}
\label{object-level analysis}
To further examine the failure cases highlighted in the Introduction—the disappearance of small, fast-moving objects and the loss of fine details under rapid motion—we compute object-level masked PSNR on the Custom dataset. We evaluate masked PSNR within manually annotated motion-blurred regions of the ball (60 frames), bat (43 frames), glove (40 frames), and pirate figure (25 frames).  As shown in ~\cref{tab:challenging_dynamic_analysis} and ~\cref{fig:analysis_quali}, our method achieves substantial PSNR gains for the ball, bat, and pirate figure, which are barely visible in competing reconstructions. It also improves the glove’s PSNR and renders its shape more clearly. Manual annotation was necessary because existing off-the-shelf models~\cite{jocher2026ultralytics,karaev2025cotracker3,carion2026sam,zhang2026megaflow} fail to reliably capture these challenging regions, as discussed above. The manually annotated masks are also provided in the supplementary material.

\begin{table}[t]
    \centering
    \caption{\textbf{Object-level masked PSNR on the Custom dataset.} Our method outperforms competing methods on all four objects, with particularly large gains for the ball, bat, and pirate figure.}
    \label{tab:challenging_dynamic_analysis}
    \footnotesize
    \setlength{\tabcolsep}{4pt}
    \begin{tabular}{@{}lcccc@{}}
        \toprule
        Method & Ball & Bat & Glove & Pirate \\
        \midrule
        RealTime4DGS~\cite{yang2024real} & 10.98 & 13.94 & 14.91 & 12.17\\
        Ex4DGS~\cite{lee2024fully} & 11.10 & 13.63 & 15.31 & 12.22\\
        4DGaussian~\cite{wu20244d} & 11.01 & 13.55 & 14.79 & 12.08\\
        Multi4D~\cite{wang2026multi4d} & 11.44 & 16.77 & 18.17 & 12.21\\
        FreeTimeGS~\cite{wang2025freetimegs} & \underline{12.41} & \underline{17.47} & \underline{21.40} & \underline{12.41} \\ 
        \midrule
        \textbf{Ours} & \textbf{19.42} & \textbf{20.68} & \textbf{22.37} & \textbf{18.93}\\
        \bottomrule
    \end{tabular}
\end{table}

\section{Conclusions}
We presented two Gaussian allocation strategies that address practical reconstruction failures in FreeTimeGS, a strong baseline for dynamic scene reconstruction: missing small, fast-moving objects and degraded details under rapid motion. \emph{Seed Injection} reallocates existing Gaussians to under-reconstructed regions using multi-view observations, enabling recovery where conventional relocation methods fail to provide sufficient support. \emph{Temporal Biased Relocation} uses time-resolved optimization signals to guide relocated Gaussians toward timestamps requiring further refinement. Together, these strategies improve the reconstruction of challenging dynamic content across multiple datasets without increasing the total Gaussian count or changing the underlying FreeTimeGS representation.

\paragraph{Limitations.}
\emph{Seed Injection} relies on frame differences and synchronized multi-view observations to identify dynamic regions and lift candidate pixels into 3D, requiring a fixed, synchronized multi-camera rig. Moreover, frame differences are sensitive to camera shake, which can introduce spurious motion signals. Objects with appearances similar to their surroundings may produce weak frame differences and thus be missed during 3D seed generation. We leave extending our approach to moving-camera capture and improving seed generation in such cases to future work.
{
    \small
    \bibliographystyle{ieeenat_fullname}
    \bibliography{main}
}
\clearpage
\hypersetup{pageanchor=false}
\maketitlesupplementary

\section{Additional Implementation Details}
This section provides additional implementation details for
\emph{Seed Injection} and \emph{Temporal Biased Relocation}.
Their experimental analyses are presented in
\cref{sec:additional_analysis}.

\subsection{Dynamic Error-based Seed Map}

\begin{figure*}[t]
    \centering
    \includegraphics[width=\textwidth]{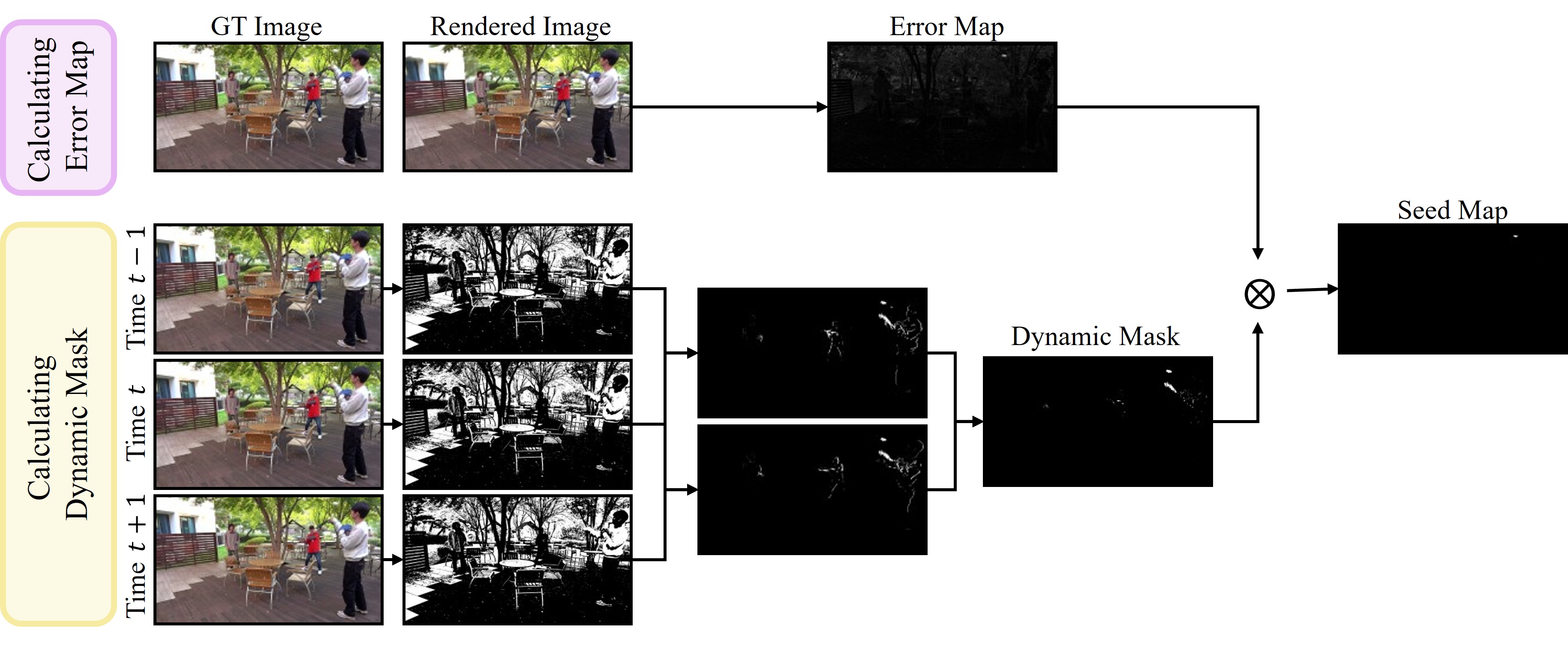}
    \caption{\textbf{Construction of the dynamic error-based seed map.} The reconstruction error map is computed from the difference between the ground-truth and rendered images.
    The dynamic mask is the intersection of two binary masks obtained by thresholding the absolute grayscale intensity differences between the current frame and each adjacent
    frame. Element-wise multiplication of the reconstruction error map and the dynamic mask yields the final seed map, highlighting under-reconstructed dynamic regions.}
    \label{fig:mask}
\end{figure*}

As illustrated in~\cref{fig:mask}, we construct the dynamic error-based seed map by combining a reconstruction error map and a binary dynamic mask to identify poorly reconstructed dynamic regions. For each camera view $v$ at timestamp $t$, we compute the pixel-wise RGB reconstruction error between the ground-truth image $I_{v,t}$ and the rendered image $\hat{I}_{v,t}$:
\begin{equation}
E_{v,t}(p)
=
\left\|
I_{v,t}(p)-\hat{I}_{v,t}(p)
\right\|_1,
\end{equation}
where $p$ denotes a pixel coordinate. This error highlights under-reconstructed regions but may also be large in static areas. To focus on dynamic content, we construct a binary mask from temporal intensity differences. We convert the current frame and its two neighboring frames into grayscale images, denoted by $I_{v,t}^{\mathrm{gray}}$, $I_{v,t-1}^{\mathrm{gray}}$, and $I_{v,t+1}^{\mathrm{gray}}$. We then threshold the absolute intensity difference between the current frame and each neighboring frame:
\begin{equation}
\begin{aligned}
B_{v,t}^{-}(p)
&=
\mathbf{1}\!\left[
\left|
I_{v,t}^{\mathrm{gray}}(p)
-
I_{v,t-1}^{\mathrm{gray}}(p)
\right| > \tau
\right], \\
B_{v,t}^{+}(p)
&=
\mathbf{1}\!\left[
\left|
I_{v,t}^{\mathrm{gray}}(p)
-
I_{v,t+1}^{\mathrm{gray}}(p)
\right| > \tau
\right],
\end{aligned}
\end{equation}
where $\mathbf{1}[\cdot]$ is the indicator function. We set $\tau=20$ for grayscale intensities in $[0,255]$. Using absolute differences captures both increases and decreases in intensity.

The dynamic mask is the intersection of the two thresholded masks:
\begin{equation}
D_{v,t}(p)
=
B_{v,t}^{-}(p)
\land
B_{v,t}^{+}(p).
\end{equation}
Thus, a pixel is retained only when its intensity differs from both neighboring frames by more than $\tau$. This suppresses changes detected against only one neighboring frame and selects regions with temporal variation relative to both neighbors.

Finally, we multiply the reconstruction error by the dynamic mask:
\begin{equation}
M_{v,t}(p)
=
D_{v,t}(p)E_{v,t}(p).
\end{equation}
The resulting seed map preserves reconstruction error magnitudes within the selected dynamic regions while suppressing errors elsewhere. Pixels with large values indicate poorly reconstructed dynamic regions.
We sample seed pixels with probabilities proportional to
${M_{v,t}(p)}$.

\begin{algorithm}[h]
\caption{Multi-view Seed Lifting}
\label{algorithm:seed_lifting}
\begin{algorithmic}[1]
\Require Candidate cross-view ray pairs $\mathcal{E}$;
ray origins $\mathbf{o}_v$, unit directions $\mathbf{d}_v^i$,
intrinsic matrices $K_v$, focal lengths $f_{x,v},f_{y,v}$,
pixel centers $(u_v^i,v_v^i)$, and seed budget $N_{\max}$
\Ensure Seed positions $\mathcal{X}_{\mathrm{seed}}$

\State $\mathcal{E}_{\mathrm{valid}} \gets \emptyset$
\For{each ray pair $(r_v^i,r_{v'}^j) \in \mathcal{E}$}
  \For{$(a,k) \in \{(v,i),(v',j)\}$}
      \State $\rho_a^k \gets
      \displaystyle
      \frac{\max(f_{x,a}^{-1},f_{y,a}^{-1})}
      {\|K_a^{-1}[u_a^k,v_a^k,1]^\top\|_2}$
  \EndFor
  \State Compute closest points
  $\mathbf{q}_v^{i,j}=\mathbf{o}_v+\lambda_{ij}^*\mathbf{d}_v^i$ and
  $\mathbf{q}_{v'}^{i,j}=\mathbf{o}_{v'}+\gamma_{ij}^*\mathbf{d}_{v'}^j$
  on the supporting lines
  \State $\delta_{ij} \gets
  \|\mathbf{q}_v^{i,j}-\mathbf{q}_{v'}^{i,j}\|_2$
  \State $\mathrm{valid} \gets
  (\lambda_{ij}^* > 0) \land (\gamma_{ij}^* > 0)
  \land \bigl(\delta_{ij} \le \lambda_{ij}^*\rho_v^i+\gamma_{ij}^*\rho_{v'}^j\bigr)$
  \If{$\mathrm{valid}$}
      \State Add $\{r_v^i,r_{v'}^j\}$ to
      $\mathcal{E}_{\mathrm{valid}}$
  \EndIf
\EndFor

\State $\mathcal{E}_{\mathrm{selected}} \gets \emptyset$
\For{each ray $r_v^i$ appearing in $\mathcal{E}_{\mathrm{valid}}$}
    \State $(v^\star,j^\star) \gets
    \arg\min\nolimits_{v',j:\,\{r_v^i,r_{v'}^j\}
    \in\mathcal{E}_{\mathrm{valid}}}\delta_{ij}$
  \State Add $\{r_v^i,r_{v^\star}^{j^\star}\}$ to
  $\mathcal{E}_{\mathrm{selected}}$
\EndFor

\State Retain at most $N_{\max}$ unique pairs from
$\mathcal{E}_{\mathrm{selected}}$ with the smallest
ray distances
\State $\mathcal{X}_{\mathrm{seed}} \gets
\left\{
\frac{\mathbf{q}_v^{i,j}+\mathbf{q}_{v'}^{i,j}}{2}
\;\middle|\;
\{r_v^i,r_{v'}^j\}\in\mathcal{E}_{\mathrm{selected}}
\right\}$
\State \Return $\mathcal{X}_{\mathrm{seed}}$
\end{algorithmic}
\end{algorithm}

\subsection{Multi-view Seed Lifting}
Following the main paper's sampling and ray-pairing
procedure, Algorithm~\ref{algorithm:seed_lifting}
details geometric validity checks and seed selection. For each ray $r_v^i$, $\mathbf{o}_v$, $\mathbf{d}_v^i$, and
$\rho_v^i$ denote its origin, unit direction, and cone
radius per unit distance; $K_v$, $(f_{x,v},f_{y,v})$,
and $(u_v^i,v_v^i)$ denote its camera intrinsic matrix,
focal lengths, and sampled pixel center.
For each candidate pair $(r_v^i,r_{v'}^j)$, we compute the
closest points on their supporting lines at signed
distances $\lambda_{ij}^*$ and $\gamma_{ij}^*$. The geometric validity check
accepts the pair if $\lambda_{ij}^*,\gamma_{ij}^*>0$ and their separation
is at most $\rho_v^i\lambda_{ij}^*+\rho_{v'}^j\gamma_{ij}^*$, the sum of the pixel-cone radii at those points.
We select each ray's nearest valid partner and use
the closest-point midpoints of at most $N_{\max}$
unique pairs with the smallest ray distances as seeds.
Here, $N_{\mathrm{seed}}$ denotes the number of pixels selected per camera frame at a given timestep, whereas $N_{\max}$ limits the total number of 3D seeds generated across all views at that timestep.

\subsection{Gaussian Injection Procedure}
The seed cache is generated once from a fixed midpoint checkpoint and reused throughout injection. The per-timestamp updates and subsequent adaptation together take total 20$N_{\mathrm{frames}}$ iterations, which are included in the original training budget.

\begin{algorithm}[H]
\caption{Gaussian Injection Procedure}
\label{algorithm:gaussian_injection}
\begin{algorithmic}[1]
\State Train FreeTimeGS~\cite{wang2025freetimegs} for half of the training budget
\State Pause optimization and cache 3D seeds for all
timestamps using the current model snapshot
\State Disable relocation
\For{$t=1,\ldots,N_{\mathrm{frames}}$}
    \State Reinitialize low-opacity Gaussians at
    cached seed positions for timestamp $t$
    \State Train for 10 iterations using observations at $t$
\EndFor
\State Resume original frame sampling and train for
$10N_{\mathrm{frames}}$ iterations with relocation disabled
\State Restore relocation and train until the original
budget is exhausted
\end{algorithmic}
\end{algorithm}

\subsection{Temporal Biased Relocation}

Algorithm~\ref{alg:temporal_relocation} summarizes
\emph{Temporal Biased Relocation}. We follow FreeTimeGS~\cite{wang2025freetimegs} for
source and target selection, while using timestamp-wise
mean 2D positional gradient magnitudes to determine
the relocation time. The duration is adjusted using
the target's temporal support boundaries, defined
at a temporal opacity of $0.1$.

\begin{algorithm}[H]
\caption{Temporal Biased Relocation}
\label{alg:temporal_relocation}
\begin{algorithmic}[1]
\Require Gaussian set $\mathcal{G}$, timestamp-wise
mean gradient magnitudes $a_{g,t}$, minimum duration
$\epsilon$

\State Select source and target $g$ following FreeTimeGS
\State $P_g(t) \gets a_{g,t}/
\sum_{u\in\mathcal{T}_g}a_{g,u}$
\State Sample $t^* \sim P_g(t)$
\State $\hat{\mu}_t \gets t^*$,
       $\hat{\mu}_x \gets \mu_x+v(t^*-\mu_t)$
\State $r \gets d\sqrt{2\ln 10}$
\State $\Delta \gets
\min\{|t^*-(\mu_t-r)|,\;|t^*-(\mu_t+r)|\}$
\State $\hat{d} \gets
\min\!\left(d,\;
\max\!\left(\epsilon,\frac{\Delta}{\sqrt{2\ln 10}}\right)
\right)$
\State Apply the remaining FreeTimeGS relocation updates
\end{algorithmic}
\end{algorithm}
\section{Additional Evaluation Details}
\subsection{Dynamic and Highly Dynamic Masks}

\begin{figure*}[t]
    \centering
        \includegraphics[
        width=1\textwidth,
        trim={1mm 0mm 0mm 1mm},
        clip
    ]{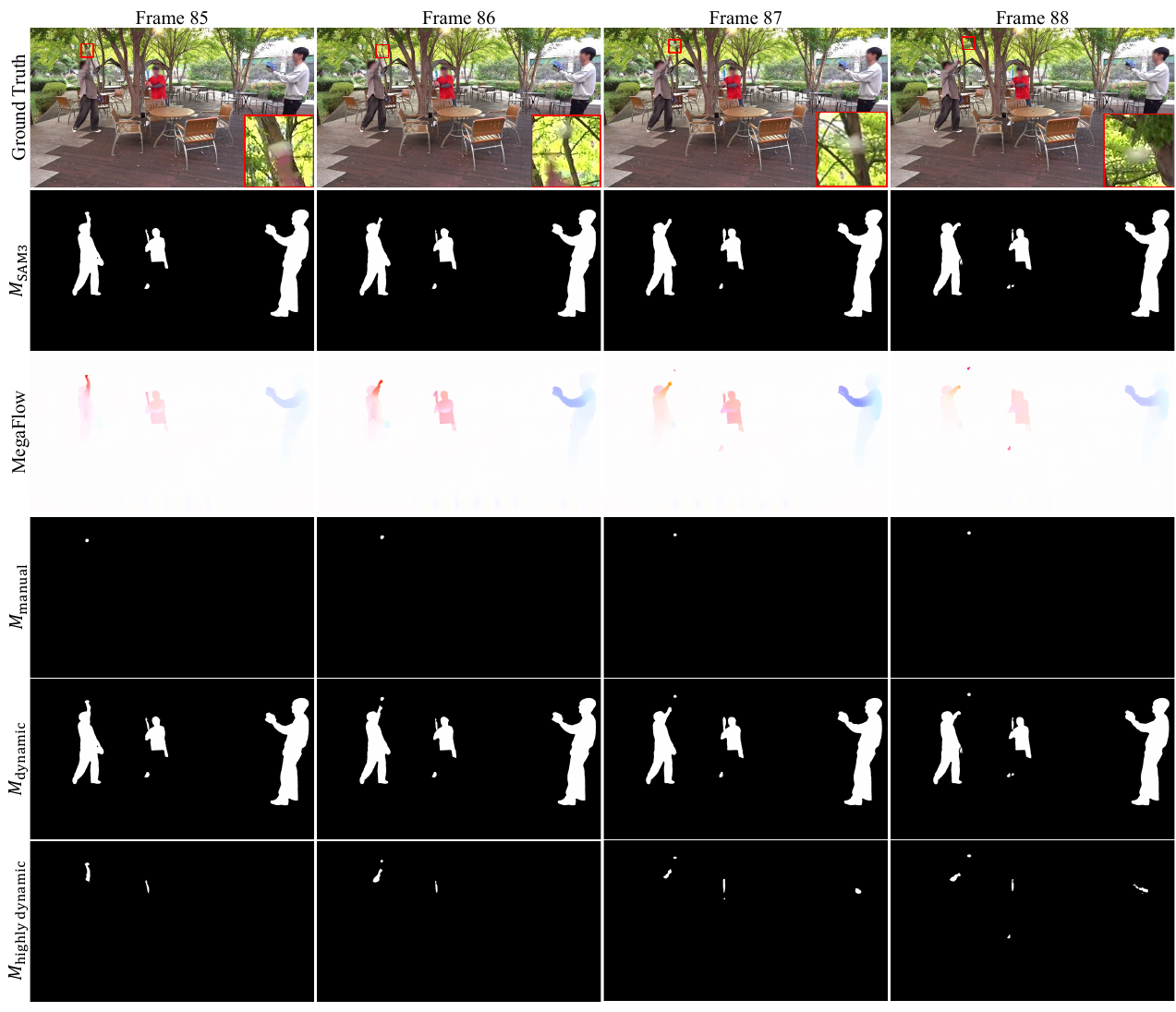}
    \caption{\textbf{Construction of dynamic and highly dynamic evaluation masks.}
    The motion-blurred ball is barely visible in the ground-truth images and is not reliably captured by SAM3 or MegaFlow. We supplement missed regions with manual annotations for CMU Panoptic and Custom.}
    \label{fig:masks}
\end{figure*}

As shown in \cref{fig:masks}, the ball is severely motion-blurred and barely visible even in the ground-truth images, making it difficult even for human observers to identify from individual frames. SAM3~\cite{carion2026sam} and MegaFlow~\cite{zhang2026megaflow} also fail to capture it reliably: the ball is either entirely missed or only partially captured, with some portions of its blurred appearance detected and others omitted. We therefore supplement the automatically generated masks with manual annotations for regions that SAM3 struggles to capture, including the bat, glove, ball, and pirate figure in Custom and the balls in CMU Panoptic~\cite{joo2015panoptic}, following the procedure described below. 

\textbf{Dynamic masks.}
We construct dynamic masks using SAM3 with text prompts and add manual annotations for regions missed by SAM3:
\[
M_{\mathrm{dynamic}}
=
M_{\mathrm{SAM3}} \cup M_{\mathrm{manual}}.
\]

\textbf{Highly dynamic masks.}
We construct highly dynamic masks by intersecting the SAM3 masks with MegaFlow-based motion masks and then adding manual annotations:
\[
M_{\mathrm{highly\ dynamic}}
=
(M_{\mathrm{SAM3}} \cap M_{\mathrm{MegaFlow}})
\cup M_{\mathrm{manual}},
\]
where $M_{\mathrm{MegaFlow}}$ is a binary mask indicating pixels whose optical flow magnitude exceeds $\theta_{\mathrm{flow}}$.
We set $\theta_{\mathrm{flow}}=5$ pixels for N3DV~\cite{li2022neural} and CMU Panoptic~\cite{joo2015panoptic}, accounting for the relatively mild motion in N3DV and the lower image resolution of CMU Panoptic. For the high-resolution ($1920\times1080$) outdoor scenes in Custom, we use $\theta_{\mathrm{flow}}=10$ pixels.
Manual annotations are used only for CMU Panoptic and Custom; no manual annotations are used for N3DV.

\subsection{Object-Level Evaluation Masks}
\begin{figure}[H]
    \centering
    \includegraphics[
        width=\columnwidth,
        trim={3mm 0mm 0mm 2.5mm},
        clip
    ]{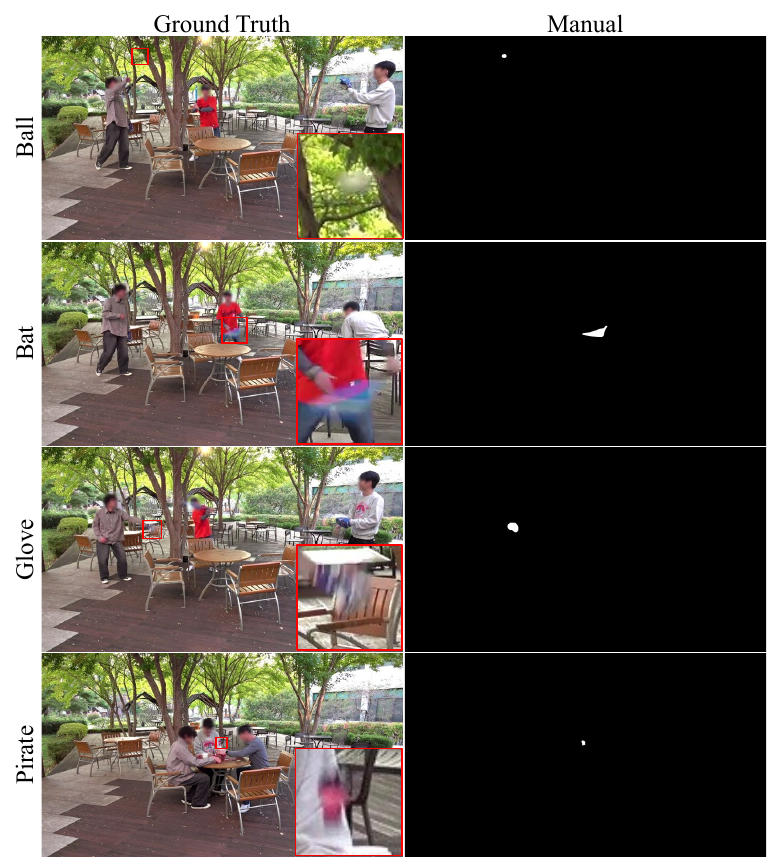}
    \caption{
       \textbf{Examples of the manually annotated masks used for object-level evaluation.}
        The masks cover motion-blurred object regions in the ground-truth images.
    }
    \label{fig:object_eval_masks}
\end{figure}

We evaluate object-level masked PSNR within manually annotated motion-blurred regions of the ball (60 frames), bat (43 frames), glove (40 frames), and pirate figure (25 frames) in the Custom dataset. For each object, we inspect the ground-truth frames from the onset of rapid motion until the motion stops and annotate only the motion-blurred object regions. As illustrated in \cref{fig:object_eval_masks}, these regions exhibit severe motion blur, making their appearance and boundaries difficult to identify.

\subsection{Full Quantitative Results}

We provide per-scene quantitative results in \cref{tab:n3dv_scene_quantitative,tab:cmu_scene_quantitative,tab:custom_scene_quantitative} for all evaluated datasets, complementing the aggregate results reported in the main paper.
In our evaluation pipeline, LPIPS~\cite{zhang2018unreasonable} is computed using AlexNet with RGB inputs in $[0,1]$, following FreeTimeGS++~\cite{lee2026freetimegs++}.
We report DSSIM-$k=(1-\mathrm{SSIM}_k)/2$, where $\mathrm{SSIM}_k$ uses a data-range parameter of $k\in\{1,2\}$~\cite{wang2004image}.
For each scene, we compute Dynamic and Highly Dynamic PSNR by aggregating the squared RGB errors over all masked pixels across the test-camera frames. The aggregated error is normalized by the total number of masked RGB values to obtain the MSE, which is then converted to PSNR.

\subsection{N3DV Qualitative Results}
\cref{fig:n3dv_qualitative} presents additional qualitative comparisons on N3DV~\cite{li2022neural}.
Our method reconstructs dynamic regions with sharper details and clearer object boundaries than Realtime4DGS~\cite{yang2024real} and FreeTimeGS~\cite{wang2025freetimegs}.
In particular, the knife in the fourth row is more clearly visible in our result, whereas competing methods struggle to preserve its thin structure.

\section{Additional Ablation Studies}

\subsection{Seed Injection}

We isolate the contribution of \emph{Seed Injection} by enabling only this module on top of the original FreeTimeGS~\cite{wang2025freetimegs} baseline, without \emph{Temporal Biased Relocation}. We conduct ablation studies on three key hyperparameters: the number of seed candidates per camera view $N_{\mathrm{seed}}$, the maximum number of injected Gaussians per timestamp $N_{\max}$, and the number of camera views $C$, while keeping the remaining training settings fixed.

\paragraph{Number of Seed Candidates.}

\begin{figure*}[h]
    \centering
    \includegraphics[width=\textwidth]{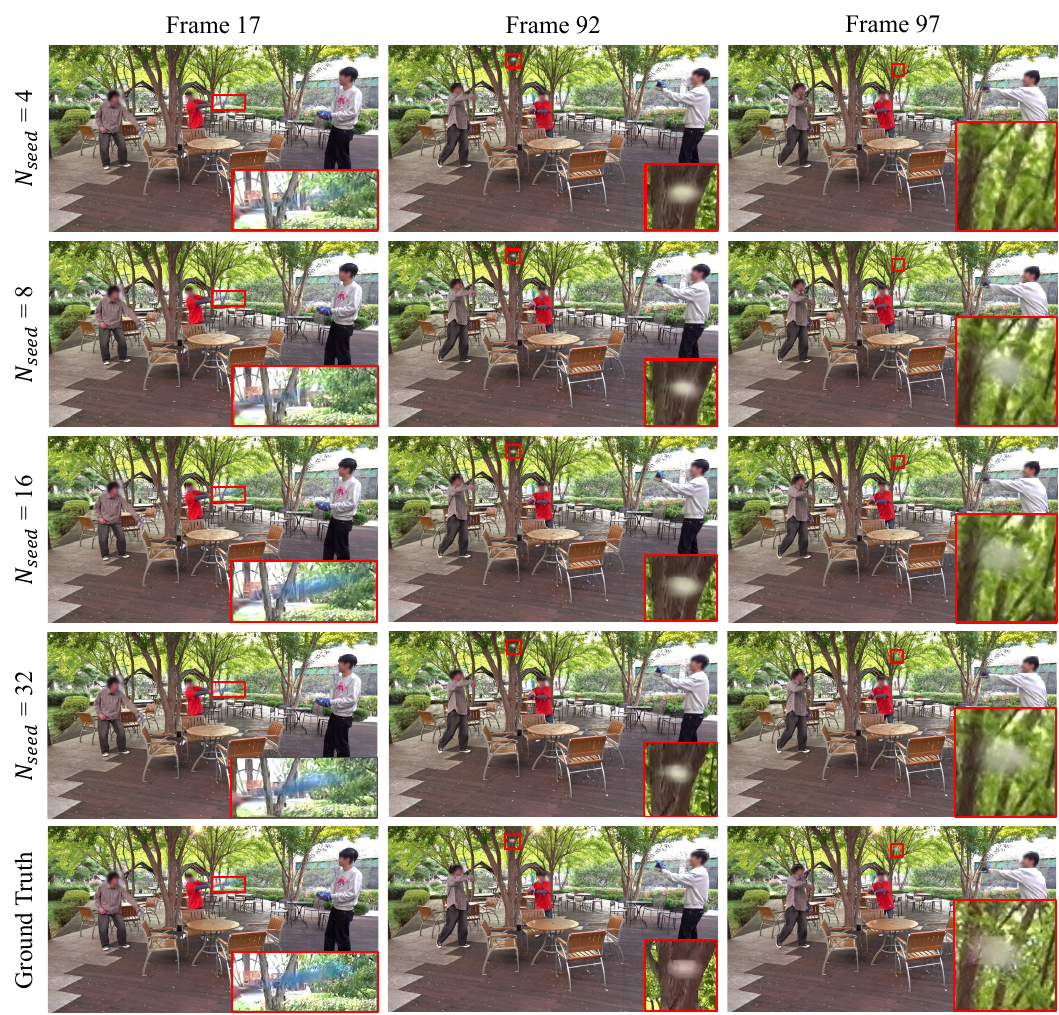}
    \caption{
    \textbf{Qualitative ablation on the number of seed candidates sampled per camera view on the Custom Baseball dataset.}
    We compare different values of $N_{\mathrm{seed}}$ across three representative frames containing small and rapidly moving objects.
    Increasing $N_{\mathrm{seed}}$ leads to more consistent recovery of challenging dynamic content, including the baseball and bat.
    }
    \label{fig:nseed_ablation}
\end{figure*}

We further investigate the effect of $N_{\mathrm{seed}}$, which controls the number of seed candidates sampled from each camera view during \emph{Seed Injection}.
The primary effect of \emph{Seed Injection} is to recover small and rapidly moving objects that may be completely missing in the baseline reconstruction and to improve fine details in fast-moving regions. Since these improvements are spatially localized to small regions and often persist for only a short temporal duration, their impact may not be fully reflected in image-level quantitative metrics (e.g., PSNR). We therefore provide a qualitative ablation with $N_{\mathrm{seed}} \in \{4,8,16,32\}$ in \cref{fig:nseed_ablation}.

As shown in \cref{fig:nseed_ablation}, the effect of $N_{\mathrm{seed}}$ varies depending on the frame and the dynamic content.
At frame 17, the baseball bat is barely visible with $N_{\mathrm{seed}}=4$ and $8$. The bat begins to emerge at $N_{\mathrm{seed}}=16$, but its structure remains unclear, while $N_{\mathrm{seed}}=32$ produces a more clearly visible reconstruction.
At frame 92, in contrast, the rapidly moving baseball is successfully reconstructed across all tested values of $N_{\mathrm{seed}}$.
At frame 97, the baseball is completely missing with $N_{\mathrm{seed}}=4$, but becomes clearly visible from $N_{\mathrm{seed}}=8$ onward. Overall, these results show that the number of seeds required for successful reconstruction can vary across frames and dynamic content. While $N_{\mathrm{seed}}=16$ is generally sufficient to recover the target objects, we use $N_{\mathrm{seed}}=32$ in our main experiments to ensure more robust and stable seed generation across different frames and scenes.

\begin{figure*}[h]
    \centering
    \includegraphics[width=\textwidth]{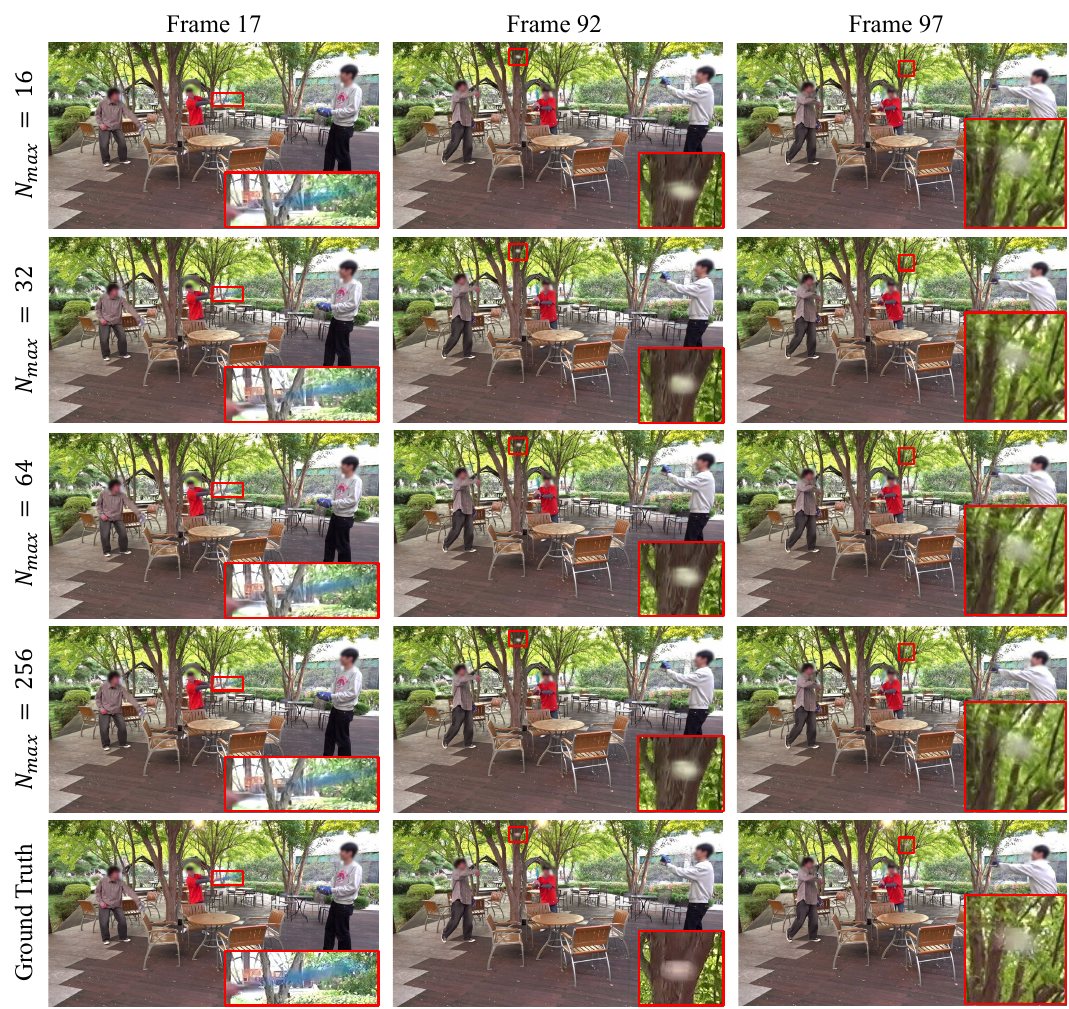}
    \caption{
   \textbf{ Qualitative ablation on the maximum number of seeds retained for Gaussian injection per timestamp on the Custom Baseball dataset.}
    We compare $N_{\max} \in \{16,32,64,256\}$ across three frames.
    Even with $N_{\max}=16$, challenging dynamic content, including the rapidly moving baseball and bat is successfully reconstructed.
    }
    \label{fig:supple_ablation_n_max}
\end{figure*}

\paragraph{Maximum Number of Injected Seeds.}
While $N_{\mathrm{seed}}$ controls the number of seed candidates sampled from each camera view, $N_{\max}$ determines how many of the resulting 3D seeds are ultimately retained for Gaussian injection per timestamp.
We therefore investigate how sensitive the reconstruction is to the final number of injected Gaussians.
As shown in \cref{fig:supple_ablation_n_max}, the reconstruction remains largely consistent across $N_{\max} \in \{16,32,64,256\}$.
Even with $N_{\max}=16$, challenging dynamic content, including the rapidly moving baseball and bat, is successfully reconstructed.
This suggests that reconstruction quality is relatively insensitive to the final number of injected Gaussians once a sufficient pool of seed candidates is available.

\begin{figure*}[h]
    \centering
    \includegraphics[width=\textwidth]{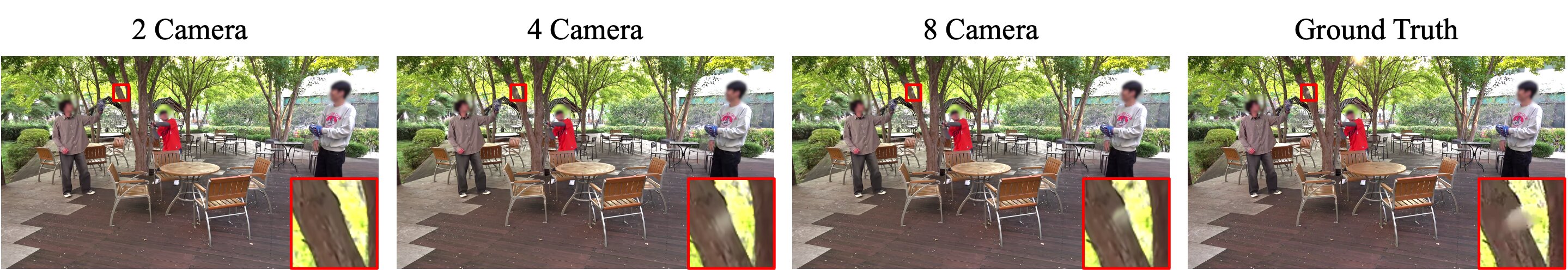}
    \caption{
    \textbf{Qualitative ablation on the number of camera views $C$ used for multi-view seed generation at frame 9 of the Custom Baseball dataset.}
    Using only two views results in inaccurate reconstruction of the rapidly moving baseball, whereas four or more views successfully recover the object with similar reconstruction quality.
    }
    \label{fig:supple_ablation_n_cam}
\end{figure*}

\paragraph{Number of Camera Views.}
Given the importance of having sufficient candidate rays for reliable seed construction, we additionally study the effect of the number of camera views $C$ used for multi-view seed generation.
\Cref{fig:supple_ablation_n_cam} shows qualitative results for $C \in \{2,4,8\}$.
With only two camera views, the limited set of cross-view ray candidates can lead to less reliable seed generation, resulting in inaccurate reconstruction of the rapidly moving baseball.
In contrast, using four or more views successfully recovers the baseball, with similar reconstruction quality for $C=4$ and $C=8$.
These results suggest that our method does not require a large number of camera views for effective seed generation, with $C=4$ already sufficient for this sequence.
Nevertheless, we use $C=8$ throughout our main experiments to provide more reliable multi-view correspondences across different frames and scenes.

\subsection{Temporal Biased Relocation}

We isolate the contribution of \emph{Temporal Biased Relocation} (TBR) by enabling only this module on top of
the original FreeTimeGS~\cite{wang2025freetimegs} baseline, without \emph{Seed Injection}. We compare the original FreeTimeGS baseline with two TBR variants (TBR-Random and TBR (Ours)). All three methods follow the same donor-selection rule but differ in how they determine the relocated Gaussian's temporal center and adjust its duration. All variants use the same low-opacity recycling criterion, opacity/scale correction, initialization, and training settings.

\paragraph{TBR-Random.}
This variant also follows the FreeTimeGS~\cite{wang2025freetimegs} donor-selection
rule, but uniformly samples the relocation timestamp from
the selected donor's observed timestamps in the current
gradient accumulation window. It then places the relocated
Gaussian along the donor's trajectory and applies the same
duration-adjustment rule as our method. This control
isolates the benefit of gradient-guided temporal selection
by retaining temporal recentering while removing the
gradient-based preference among timestamps.

\paragraph{TBR (Ours).}
Our method uses gradient-guided sampling to select the relocated Gaussian's temporal center within the gradient window. Unlike TBR-Random, it favors candidate times according to the gradient signal rather than sampling without this guidance. The comparison therefore evaluates whether gradient-based temporal selection provides a more effective allocation of the fixed Gaussian budget.

\begin{figure}[h]
    \centering
    \includegraphics[width=1\columnwidth]{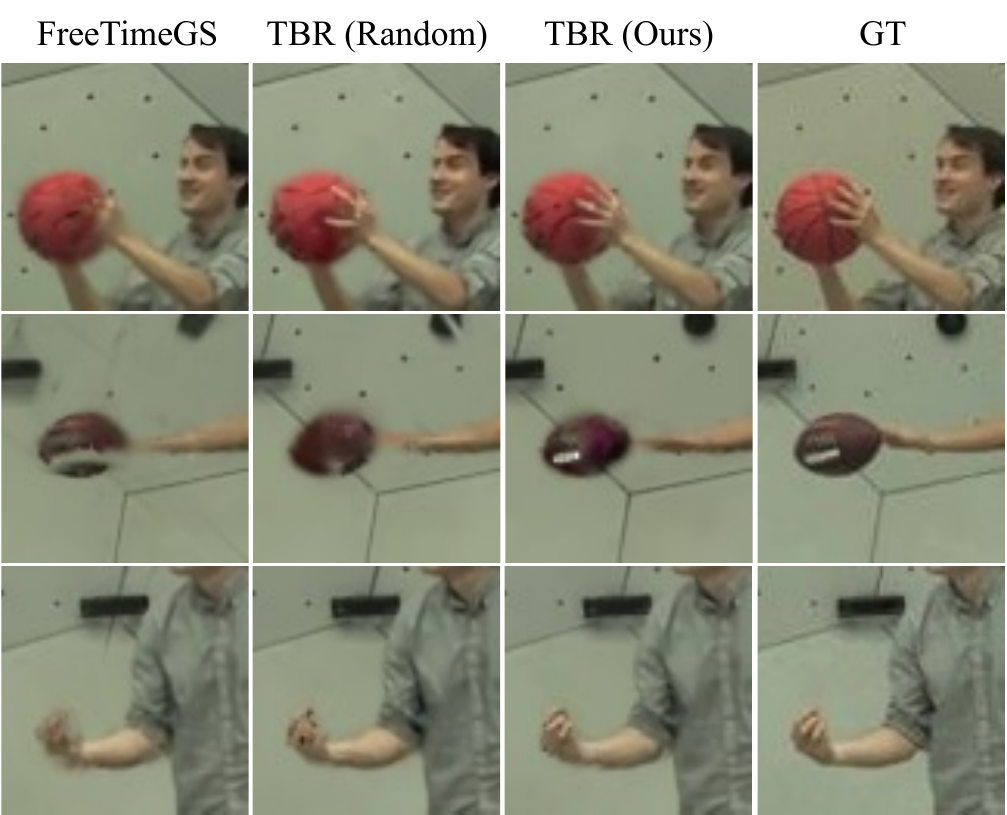}
    \caption{\textbf{Qualitative comparison of Temporal Biased Relocation variants on CMU Panoptic} Columns show FreeTimeGS, TBR-Random, our gradient-guided TBR, and ground truth. The Basketball, Football, and Juggle examples highlight differences in finger structure, hand contours, and manipulated objects. Our method preserves these details more clearly than the alternative temporal-center selection strategies.}
    \label{fig:cmu_tbr_ablation}
\end{figure}

\cref{tab:cmu_tbr_ablation} shows that our gradient-guided TBR achieves higher Dynamic and Highly Dynamic PSNR than both the original FreeTimeGS~\cite{wang2025freetimegs} baseline and TBR-Random, indicating improved reconstruction of moving regions. Figure~\ref{fig:cmu_tbr_ablation} illustrates these improvements in selected examples: our method better preserves finger structure, hand contours, and manipulated objects, whereas the alternative strategies exhibit blurred or displaced boundaries.

\begin{table}[h]
    \centering
    \caption{Comparison of TBR variants on the CMU Juggle, Football, and Basketball sequences.}
    \label{tab:cmu_tbr_ablation}
    \setlength{\tabcolsep}{3pt}
    \resizebox{\linewidth}{!}{%
    \begin{tabular}{lccDH}
        \toprule
        Method
        & PSNR $\uparrow$
        & LPIPS $\downarrow$
        & \makecell{Dynamic\\PSNR $\uparrow$}
        & \makecell{Highly Dynamic\\PSNR $\uparrow$} \\
        \midrule
        FreeTimeGS~\cite{wang2025freetimegs}
        & 28.22 & 0.083
        & 26.24 & 23.26 \\
        FreeTimeGS w/ TBR Random
        & \underline{28.31} & \underline{0.058}
        & 26.41 & \underline{23.58} \\
        \midrule
        \textbf{FreeTimeGS w/ TBR (Ours)}
        & \textbf{28.57} & \textbf{0.056}
        & \textbf{26.89} & \textbf{24.23} \\
        \bottomrule
    \end{tabular}%
    }
\end{table}

\section{Additional Analysis}
\label{sec:additional_analysis}

Unless otherwise specified, all analyses of \emph{Seed Injection} (SI) and \emph{Temporal Biased Relocation} (TBR) are conducted by independently enabling each component on top of the original FreeTimeGS baseline (i.e., FreeTimeGS w/ SI or FreeTimeGS w/ TBR), without combining the two.

\subsection{Seed Injection}

\paragraph{Ray-geometry consistency.}
While the dynamic error-based seed map guides pixel sampling toward under-reconstructed dynamic regions, seed positions are determined by multi-view ray geometry. Specifically, each seed is placed at the midpoint of the closest points on an accepted cross-view ray pair, analogous to two-view triangulation. To assess consistency with the source observations, we project each seed back into its two generating camera views and measure the Euclidean distance to the corresponding source pixel centers.

Using the frozen checkpoint after 50K warm-up iterations, we repeat seed generation with the original settings across all 250 frames of the Baseball sequence, without additional optimization. This produces 35,422 seeds and 70,844 source-view observations, as each seed is generated from a pair of observations. The median and 90th-percentile reprojection errors across these observations are $0.343$ and $0.846$ pixels, respectively. Furthermore, $93.21\%$ of seeds have reprojection errors within one pixel in both source views.

These results show that the generated seeds are geometrically consistent with their source observations, with subpixel reprojection errors for most observations. This is consistent with our geometric validity criterion, which requires positive signed distances ($\lambda_{ij}^*,\gamma_{ij}^*>0$) and constrains the distance between the closest points to be no greater than $\rho_v^i\lambda_{ij}^*+\rho_{v'}^j\gamma_{ij}^*$, the sum of the corresponding pixel-cone radii.

\begin{figure}[h]
    \centering
    \includegraphics[width=0.95\columnwidth, trim=7 0 0 0, clip]{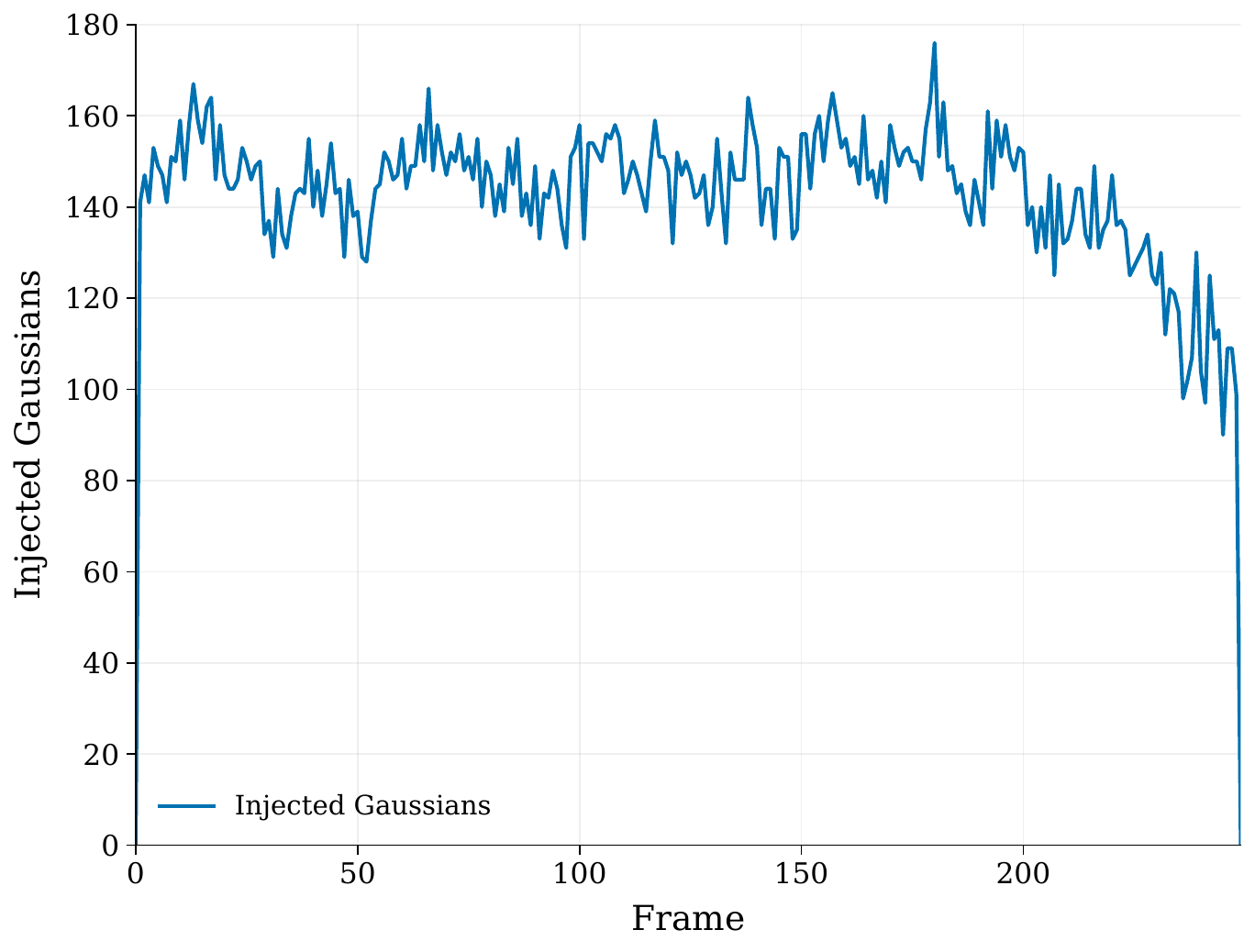}
    \caption{\textbf{Number of Gaussians injected per timestamp in the Custom Baseball sequence.} The number of injected Gaussians remains below the maximum of 256.}
    \label{fig:injected_gs}
\end{figure}

\cref{fig:injected_gs} shows that an average of 142 Gaussians were injected per timestamp during the original Custom Baseball training run. In the seed-generation replay described above, 65,260 of 7,110,656 cross-view ray pairs passed the geometric validity checks. Nearest-partner selection retained 35,422 unique pairs, all within the per-timestamp seed budget of 256.

\begin{figure*}[!t]
    \centering
    \includegraphics[width=1\linewidth]{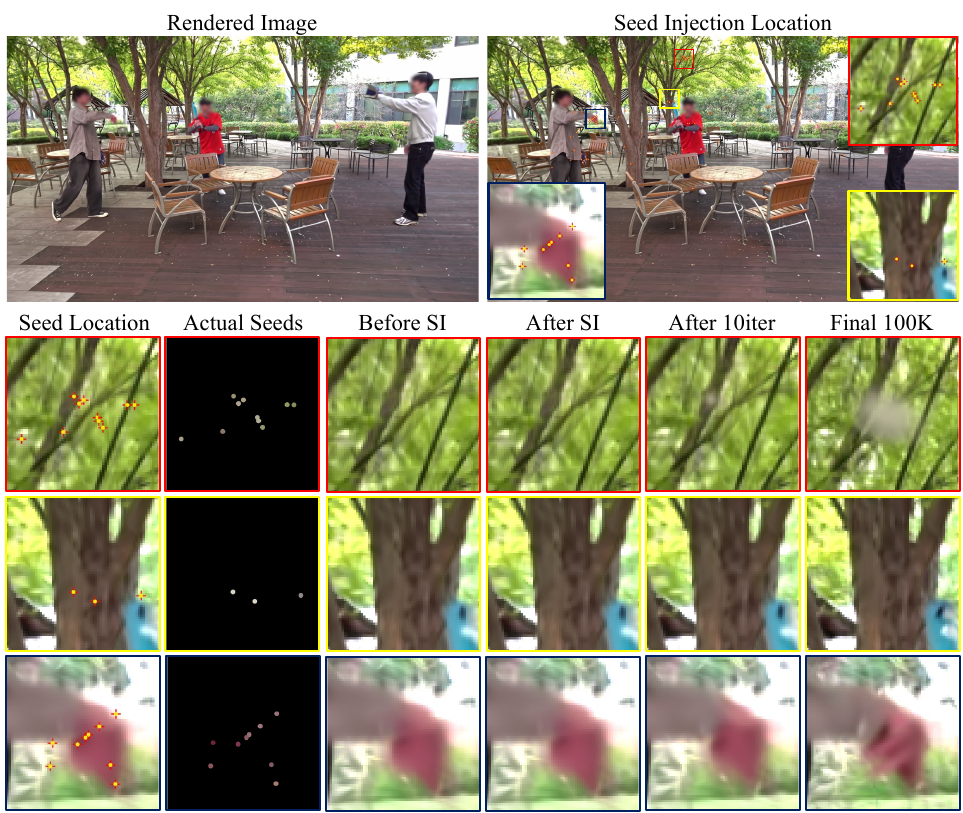}
    \caption{\textbf{Seed placement and optimization after Seed Injection on Custom Baseball.}
    \textbf{Top:} Rendered image and projected seed locations, with colored boxes indicating the regions enlarged below.
    \textbf{Bottom:} Projected seed locations, injected Gaussians rendered in isolation, and reconstructions before injection, immediately after injection, after 10 local optimization steps, and at the end of 100K training iterations. The injected Gaussians initially have small scales and low opacity, resulting in limited visual impact even for potentially inaccurate placements. Subsequent optimization strengthens useful contributions while suppressing unnecessary ones, progressively recovering the target content without noticeable persistent floating artifacts. As illustrated in the first row of the bottom panel, the injected Gaussians gradually grow to reconstruct the baseball, which becomes increasingly visible throughout optimization.
    }
    \label{fig:seed_evolution}
\end{figure*}

\begin{figure*}[t]
    \centering
    \includegraphics[width=1\linewidth]{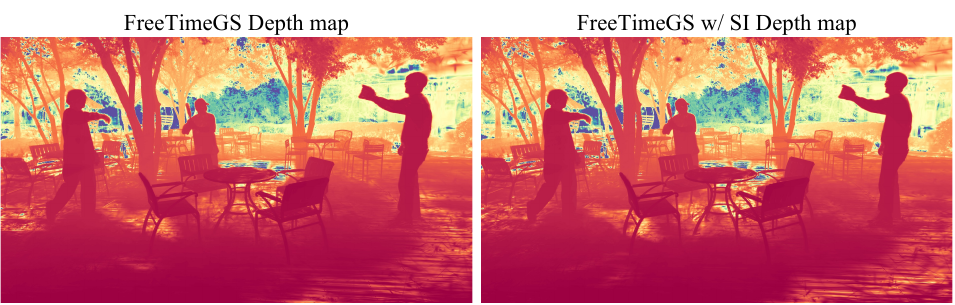}
    \caption{\textbf{Depth-map comparison on Custom Baseball.}
    Rendered depth maps from FreeTimeGS (left) and FreeTimeGS with \emph{Seed Injection} (right).
    Seed Injection preserves the overall scene depth structure without introducing noticeable floating artifacts, while successfully reconstructing the ball in the upper-middle region, as indicated by its distinct depth.}
    \label{fig:depth}
\end{figure*}

\paragraph{Floating Artifact Analysis.}
Although our geometric validity checks constrain seed placement, imperfect cross-view correspondences may still introduce seeds at inaccurate 3D locations. To investigate whether such seeds produce floating artifacts, \cref{fig:seed_evolution} visualizes the projected seed locations, the injected Gaussians rendered in isolation, and the evolution of the reconstruction throughout optimization.

As shown in the figure, the injected Gaussians initially have very small scales and low opacity, limiting their visual impact. While some seeds exhibit colors consistent with the surrounding scene content, others show apparent mismatches, suggesting that not all seeds are placed at accurate 3D locations. Nevertheless, subsequent optimization can increase the contributions of useful Gaussians while suppressing those that do not explain the observations.

The bottom panel of \cref{fig:seed_evolution} further illustrates this behavior. In the first row, the baseball is barely visible immediately after injection and remains difficult to distinguish after 10 local optimization steps, with only a faint floater-like artifact appearing in front of the tree. In contrast, the other rows show potentially inaccurate seed placements, such as white-colored Gaussians injected near the tree trunk in the second row and skin-colored Gaussians placed outside the hand region in the third row. Despite these mismatches, their visual contributions diminish during subsequent optimization, leaving no noticeable persistent floating artifacts in the final reconstructions.

These observations highlight that \emph{Seed Injection} does not require every seed to be accurately placed initially. Instead, it provides candidate Gaussians in potentially under-reconstructed regions, allowing subsequent optimization to strengthen useful contributions while suppressing unnecessary ones. Together with the small-scale and low-opacity initialization, this enables the method to tolerate imperfect seed placements without introducing noticeable floating artifacts in the visualized examples.

The corresponding depth maps~\cref{fig:depth} show no noticeable floating artifacts in the visualized examples. These observations suggest that \emph{Seed Injection} avoids introducing noticeable artifacts through small initial scales, low initial opacity, and subsequent optimization.

\begin{figure*}[!t]
    \centering
    \includegraphics[width=\linewidth]{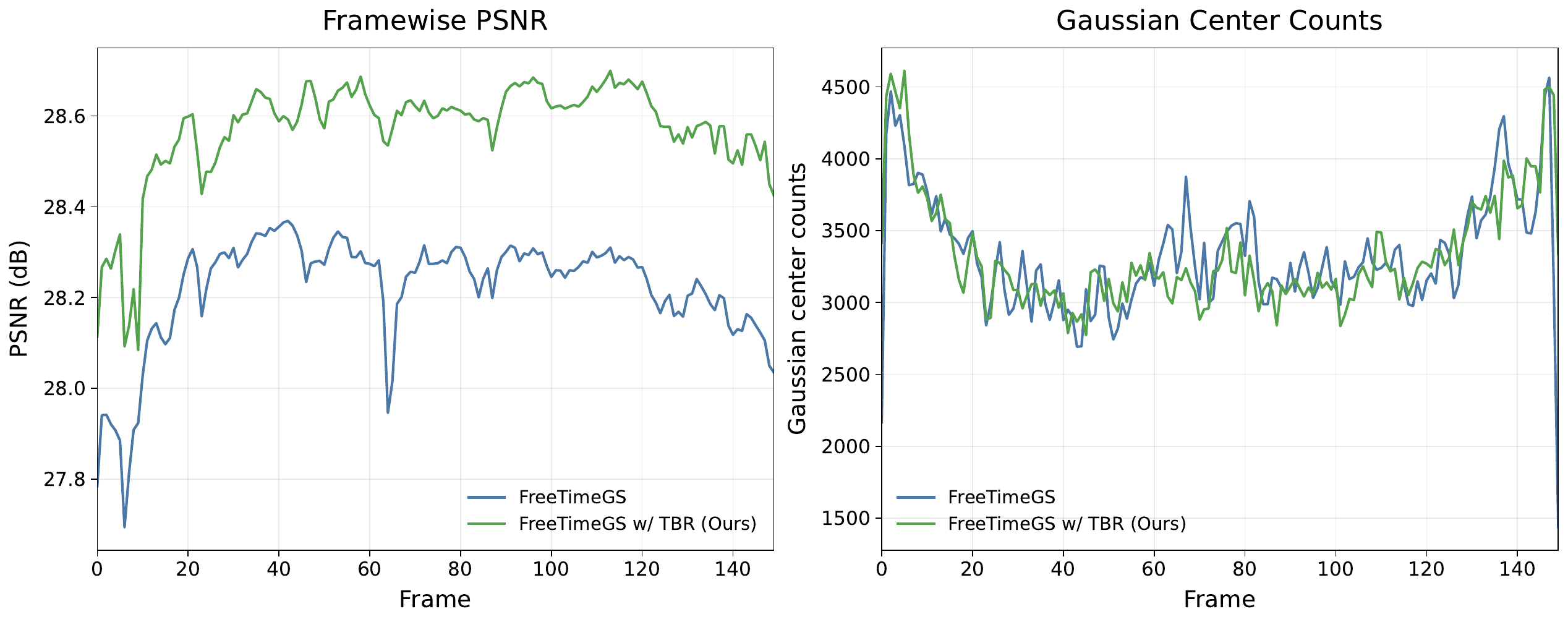}
    \caption{\textbf{Analysis of Temporal Biased Relocation on CMU Panoptic.}
    Frame-wise PSNR (left) and the distribution of Gaussian temporal centers (right), averaged over three highly dynamic sequences: Basketball, Football, and Juggle. Our TBR consistently improves reconstruction quality across timestamps while maintaining a temporal center distribution similar to that of FreeTimeGS.}
    \label{fig:tbr_analysis}
\end{figure*}

\begin{figure*}[!t]
    \centering
    \includegraphics[width=1\linewidth]{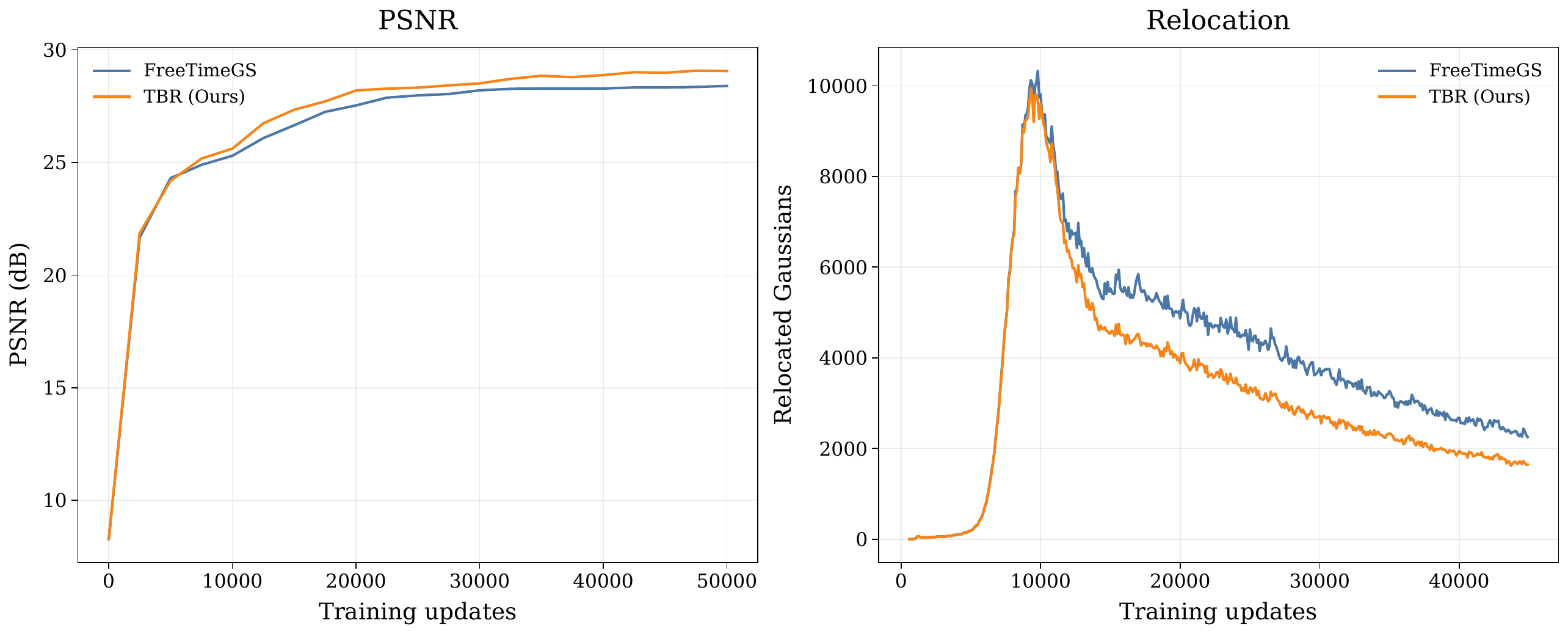}
    \caption{\textbf{Training behavior of Temporal Biased Relocation.} PSNR (left) and the number of relocated Gaussians (right) over training iterations, averaged over the CMU PanopticBasketball, Football, and Juggle sequences. Compared with FreeTimeGS, TBR achieves higher PSNR while relocating fewer Gaussians during the later stages of training.}
    \label{fig:tbr_training}
\end{figure*}

\subsection{Temporal Biased Relocation}
\paragraph{Temporal Distribution Analysis.}
We further analyze the effect of \emph{Temporal Biased Relocation} (TBR) on three highly dynamic CMU Panoptic~\cite{joo2015panoptic} sequences: Basketball, Football, and Juggle. As shown in \cref{fig:tbr_analysis} (right), the distribution of Gaussian temporal centers remains largely similar to that of the original FreeTimeGS~\cite{wang2025freetimegs}, indicating that TBR does not simply concentrate Gaussians at highly dynamic timestamps. Furthermore, the frame-wise PSNR comparison (left) shows consistent improvements across nearly all timestamps, rather than improvements in highly dynamic regions at the expense of relatively static ones.

These observations suggest that the gains from TBR do not primarily arise from reallocating Gaussians away from well-reconstructed timestamps toward highly dynamic ones. Instead, TBR provides more effective temporal relocation guidance, allowing optimization to reach a better reconstruction solution while largely preserving the overall temporal distribution of Gaussians.

\paragraph{Training Behavior.}
We further analyze the training behavior of TBR on three highly dynamic CMU Panoptic sequences: Basketball, Football, and Juggle. As shown in \cref{fig:tbr_training}, TBR achieves higher PSNR than the original FreeTimeGS throughout most of training, suggesting faster convergence toward a better reconstruction solution. Meanwhile, the number of relocated Gaussians is consistently lower during the later stages of training. This suggests that gradient-guided temporal relocation helps Gaussians reach more suitable spatiotemporal locations earlier, reducing the need for repeated relocation. Together, these observations indicate that TBR improves the efficiency of Gaussian allocation, achieving better reconstruction quality with fewer relocation operations.

\begin{table*}[t]
    \centering
    \footnotesize
    \caption{Scene-wise quantitative comparison on N3DV.}
    \label{tab:n3dv_scene_quantitative}
    \setlength{\tabcolsep}{3pt}
    \renewcommand{\arraystretch}{1.12}
    \resizebox{\textwidth}{!}{%
    \begin{tabular}{@{}l*{7}{cc}@{}}
        \toprule
        \multicolumn{15}{c}{\textbf{N3DV}} \\
        \midrule
        & \multicolumn{2}{c}{Coffee Martini} & \multicolumn{2}{c}{Cook Spinach} & \multicolumn{2}{c}{Cut Roasted Beef} & \multicolumn{2}{c}{Flame Salmon} & \multicolumn{2}{c}{Flame Steak} & \multicolumn{2}{c}{Sear Steak} & \multicolumn{2}{c}{Avg.} \\
        \cmidrule(lr){2-3}\cmidrule(lr){4-5}\cmidrule(lr){6-7}\cmidrule(lr){8-9}\cmidrule(lr){10-11}\cmidrule(lr){12-13}\cmidrule(l){14-15}

        Method & PSNR$\uparrow$ & LPIPS$\downarrow$ & PSNR$\uparrow$ & LPIPS$\downarrow$ & PSNR$\uparrow$ & LPIPS$\downarrow$ & PSNR$\uparrow$ & LPIPS$\downarrow$ & PSNR$\uparrow$ & LPIPS$\downarrow$ & PSNR$\uparrow$ & LPIPS$\downarrow$ & PSNR$\uparrow$ & LPIPS$\downarrow$ \\
        \midrule
        RealTime4DGS~\cite{yang2024real}
        & 28.12 & 0.125
        & 33.23 & 0.088
        & 33.40 & 0.086
        & 28.97 & 0.114
        & \underline{34.11} & 0.065
        & 33.71 & 0.071
        & 31.92 & 0.091 \\
        Ex4DGS~\cite{lee2024fully}
        & 27.37 & 0.118
        & 30.05 & 0.087
        & 33.43 & 0.078
        & 29.37 & 0.100
        & 31.63 & 0.069
        & 32.66 & 0.070
        & 30.75 & 0.087 \\
        4DGaussian~\cite{wu20244d}
        & 27.14 & 0.123
        & 32.53 & 0.091
        & 32.98 & 0.097
        & 28.80 & 0.106
        & 31.90 & 0.081
        & 31.40 & 0.085
        & 30.79 & 0.097 \\
        Multi4D~\cite{wang2026multi4d}
        & 28.61 & 0.121
        & 31.47 & 0.095
        & 33.04 & 0.087
        & 29.28 & 0.112
        & 33.04 & 0.078
        & 33.33 & 0.081
        & 31.46 & 0.096 \\
        FreeTimeGS~\cite{wang2025freetimegs}
        & \underline{30.02} & \underline{0.060}
        & \underline{33.52} & \underline{0.036}
        & \underline{33.99} & \underline{0.036}
        & \underline{29.80} & \underline{0.058}
        & 33.41 & \underline{0.031}
        & \underline{34.12} & \underline{0.030}
        & \underline{32.48} & \underline{0.042} \\
        \midrule
        \textbf{Ours}
        & \textbf{31.06} & \textbf{0.058}
        & \textbf{33.78} & \textbf{0.033}
        & \textbf{34.05} & \textbf{0.033}
        & \textbf{30.73} & \textbf{0.055}
        & \textbf{34.26} & \textbf{0.026}
        & \textbf{34.18} & \textbf{0.026}
        & \textbf{33.01} & \textbf{0.039} \\
        \midrule

        Method & \DynVal{\makecell{Dynamic\\PSNR$\uparrow$}} & \HighVal{\makecell{Highly Dynamic\\PSNR$\uparrow$}} & \DynVal{\makecell{Dynamic\\PSNR$\uparrow$}} & \HighVal{\makecell{Highly Dynamic\\PSNR$\uparrow$}} & \DynVal{\makecell{Dynamic\\PSNR$\uparrow$}} & \HighVal{\makecell{Highly Dynamic\\PSNR$\uparrow$}} & \DynVal{\makecell{Dynamic\\PSNR$\uparrow$}} & \HighVal{\makecell{Highly Dynamic\\PSNR$\uparrow$}} & \DynVal{\makecell{Dynamic\\PSNR$\uparrow$}} & \HighVal{\makecell{Highly Dynamic\\PSNR$\uparrow$}} & \DynVal{\makecell{Dynamic\\PSNR$\uparrow$}} & \HighVal{\makecell{Highly Dynamic\\PSNR$\uparrow$}} & \DynVal{\makecell{Dynamic\\PSNR$\uparrow$}} & \HighVal{\makecell{Highly Dynamic\\PSNR$\uparrow$}} \\
        \midrule
        RealTime4DGS~\cite{yang2024real}
        & \DynVal{\underline{32.57}} & \HighVal{22.80}
        & \DynVal{\textbf{31.22}} & \HighVal{\underline{23.30}}
        & \DynVal{\textbf{33.93}} & \HighVal{\textbf{32.08}}
        & \DynVal{\underline{29.82}} & \HighVal{13.29}
        & \DynVal{\textbf{30.99}} & \HighVal{20.08}
        & \DynVal{33.89} & \HighVal{\textbf{25.97}}
        & \DynVal{\underline{32.07}} & \HighVal{\underline{22.92}} \\
        Ex4DGS~\cite{lee2024fully}
        & \DynVal{29.64} & \HighVal{24.23}
        & \DynVal{24.80} & \HighVal{22.72}
        & \DynVal{32.71} & \HighVal{29.17}
        & \DynVal{29.04} & \HighVal{13.54}
        & \DynVal{29.32} & \HighVal{19.79}
        & \DynVal{33.71} & \HighVal{23.15}
        & \DynVal{29.87} & \HighVal{22.10} \\
        4DGaussian~\cite{wu20244d}
        & \DynVal{26.77} & \HighVal{19.74}
        & \DynVal{28.51} & \HighVal{21.39}
        & \DynVal{30.36} & \HighVal{25.32}
        & \DynVal{26.41} & \HighVal{10.05}
        & \DynVal{28.15} & \HighVal{18.14}
        & \DynVal{32.60} & \HighVal{20.56}
        & \DynVal{28.80} & \HighVal{19.20} \\
        Multi4D~\cite{wang2026multi4d}
        & \DynVal{31.60} & \HighVal{22.79}
        & \DynVal{29.37} & \HighVal{22.01}
        & \DynVal{32.96} & \HighVal{29.83}
        & \DynVal{28.94} & \HighVal{13.34}
        & \DynVal{26.95} & \HighVal{19.29}
        & \DynVal{33.00} & \HighVal{20.24}
        & \DynVal{30.47} & \HighVal{21.25} \\
        FreeTimeGS~\cite{wang2025freetimegs}
        & \DynVal{32.46} & \HighVal{\underline{24.24}}
        & \DynVal{29.52} & \HighVal{22.73}
        & \DynVal{33.14} & \HighVal{29.64}
        & \DynVal{29.41} & \HighVal{\underline{13.59}}
        & \DynVal{30.05} & \HighVal{\underline{21.07}}
        & \DynVal{\underline{34.06}} & \HighVal{22.65}
        & \DynVal{31.44} & \HighVal{22.32} \\
        \midrule
        \textbf{Ours}
        & \DynVal{\textbf{33.03}} & \HighVal{\textbf{24.80}}
        & \DynVal{\underline{30.49}} & \HighVal{\textbf{23.40}}
        & \DynVal{\underline{33.80}} & \HighVal{\underline{30.92}}
        & \DynVal{\textbf{30.26}} & \HighVal{\textbf{15.17}}
        & \DynVal{\underline{30.80}} & \HighVal{\textbf{22.19}}
        & \DynVal{\textbf{34.64}} & \HighVal{\underline{24.10}}
        & \DynVal{\textbf{32.17}} & \HighVal{\textbf{23.43}} \\
        \midrule

        Method & DSSIM-1$\downarrow$ & DSSIM-2$\downarrow$ & DSSIM-1$\downarrow$ & DSSIM-2$\downarrow$ & DSSIM-1$\downarrow$ & DSSIM-2$\downarrow$ & DSSIM-1$\downarrow$ & DSSIM-2$\downarrow$ & DSSIM-1$\downarrow$ & DSSIM-2$\downarrow$ & DSSIM-1$\downarrow$ & DSSIM-2$\downarrow$ & DSSIM-1$\downarrow$ & DSSIM-2$\downarrow$ \\
        \midrule
        RealTime4DGS~\cite{yang2024real}
        & 0.045 & 0.025
        & 0.022 & 0.011
        & 0.021 & 0.010
        & 0.041 & 0.023
        & 0.019 & 0.009
        & 0.019 & 0.009
        & 0.028 & 0.015 \\
        
        Ex4DGS~\cite{lee2024fully}
        & 0.048 & 0.027
        & 0.029 & 0.016
        & 0.023 & 0.011
        & 0.040 & 0.022
        & 0.022 & 0.011
        & 0.021 & 0.010
        & 0.031 & 0.016 \\
        
        4DGaussian~\cite{wu20244d}
        & 0.048 & 0.027
        & 0.026 & 0.013
        & 0.026 & 0.013
        & 0.042 & 0.023
        & 0.022 & 0.011
        & 0.022 & 0.011
        & 0.031 & 0.016 \\
        
        Multi4D~\cite{wang2026multi4d}
        & 0.038 & \underline{0.021}
        & 0.021 & \underline{0.010}
        & 0.019 & \underline{0.009}
        & 0.035 & 0.019
        & 0.018 & 0.009
        & 0.017 & 0.008
        & 0.025 & \underline{0.013} \\
        
        FreeTimeGS~\cite{wang2025freetimegs}
        & \underline{0.032} & \textbf{0.017}
        & \underline{0.018} & \textbf{0.009}
        & \underline{0.018} & \underline{0.009}
        & \underline{0.031} & \underline{0.017}
        & \underline{0.016} & \underline{0.008}
        & \underline{0.015} & \underline{0.007}
        & \underline{0.022} & \textbf{0.011} \\
        \midrule
        \textbf{Ours}
        & \textbf{0.031} & \textbf{0.017}
        & \textbf{0.017} & \textbf{0.009}
        & \textbf{0.016} & \textbf{0.008}
        & \textbf{0.030} & \textbf{0.016}
        & \textbf{0.015} & \textbf{0.007}
        & \textbf{0.014} & \textbf{0.006}
        & \textbf{0.021} & \textbf{0.011} \\

        \bottomrule
    \end{tabular}%
    }
\end{table*}

\begin{table*}[t]
    \centering
    \footnotesize
    \caption{Scene-wise quantitative comparison on CMU Panoptic.}
    \label{tab:cmu_scene_quantitative}
    \setlength{\tabcolsep}{3pt}
    \renewcommand{\arraystretch}{1.12}
    \resizebox{\textwidth}{!}{%
    \begin{tabular}{@{}l*{7}{cc}@{}}
        \toprule
        \multicolumn{15}{c}{\textbf{CMU Panoptic}} \\
        \midrule

        & \multicolumn{2}{c}{Basketball}
        & \multicolumn{2}{c}{Boxes}
        & \multicolumn{2}{c}{Football}
        & \multicolumn{2}{c}{Juggle}
        & \multicolumn{2}{c}{Softball}
        & \multicolumn{2}{c}{Tennis}
        & \multicolumn{2}{c}{Avg.} \\
        \cmidrule(lr){2-3}
        \cmidrule(lr){4-5}
        \cmidrule(lr){6-7}
        \cmidrule(lr){8-9}
        \cmidrule(lr){10-11}
        \cmidrule(lr){12-13}
        \cmidrule(l){14-15}

        Method
        & PSNR$\uparrow$ & LPIPS$\downarrow$
        & PSNR$\uparrow$ & LPIPS$\downarrow$
        & PSNR$\uparrow$ & LPIPS$\downarrow$
        & PSNR$\uparrow$ & LPIPS$\downarrow$
        & PSNR$\uparrow$ & LPIPS$\downarrow$
        & PSNR$\uparrow$ & LPIPS$\downarrow$
        & PSNR$\uparrow$ & LPIPS$\downarrow$ \\
        \midrule

        RealTime4DGS~\cite{yang2024real}
        & 27.67 & 0.115
        & 27.61 & 0.115
        & 27.44 & 0.109
        & 28.40 & 0.103
        & \underline{28.04} & 0.111
        & 28.05 & 0.107
        & 27.87 & 0.110 \\

        Ex4DGS~\cite{lee2024fully}
        & 24.69 & 0.195
        & 27.13 & 0.177
        & 24.82 & 0.197
        & 25.67 & 0.171
        & 24.98 & 0.171
        & 20.00 & 0.246
        & 24.55 & 0.193 \\

        4DGaussian~\cite{wu20244d}
        & 26.18 & 0.199
        & 25.51 & 0.197
        & 26.54 & 0.189
        & 26.77 & 0.183
        & 25.34 & 0.203
        & 26.77 & 0.176
        & 26.18 & 0.191 \\

        Multi4D~\cite{wang2026multi4d}
        & 26.52 & 0.151
        & 26.68 & 0.140
        & 26.21 & 0.166
        & 26.05 & 0.165
        & 27.19 & 0.145
        & 26.86 & 0.149
        & 26.58 & 0.153 \\

        FreeTimeGS~\cite{wang2025freetimegs}
        & \underline{28.01} & \underline{0.089}
        & \textbf{29.47} & \underline{0.109}
        & \underline{28.30} & \underline{0.082}
        & \underline{28.94} & \underline{0.070}
        & 27.93 & \underline{0.097}
        & \underline{28.63} & \underline{0.081}
        & \underline{28.55} & \underline{0.089} \\

        \midrule
        \textbf{Ours}
        & \textbf{28.81} & \textbf{0.073}
        & \underline{29.00} & \textbf{0.058}
        & \textbf{29.13} & \textbf{0.061}
        & \textbf{29.01} & \textbf{0.051}
        & \textbf{29.10} & \textbf{0.057}
        & \textbf{29.30} & \textbf{0.053}
        & \textbf{29.05} & \textbf{0.059} \\

        \midrule

        Method
        & \DynVal{\makecell{Dynamic\\PSNR$\uparrow$}}
        & \HighVal{\makecell{Highly Dynamic\\PSNR$\uparrow$}}
        & \DynVal{\makecell{Dynamic\\PSNR$\uparrow$}}
        & \HighVal{\makecell{Highly Dynamic\\PSNR$\uparrow$}}
        & \DynVal{\makecell{Dynamic\\PSNR$\uparrow$}}
        & \HighVal{\makecell{Highly Dynamic\\PSNR$\uparrow$}}
        & \DynVal{\makecell{Dynamic\\PSNR$\uparrow$}}
        & \HighVal{\makecell{Highly Dynamic\\PSNR$\uparrow$}}
        & \DynVal{\makecell{Dynamic\\PSNR$\uparrow$}}
        & \HighVal{\makecell{Highly Dynamic\\PSNR$\uparrow$}}
        & \DynVal{\makecell{Dynamic\\PSNR$\uparrow$}}
        & \HighVal{\makecell{Highly Dynamic\\PSNR$\uparrow$}}
        & \DynVal{\makecell{Dynamic\\PSNR$\uparrow$}}
        & \HighVal{\makecell{Highly Dynamic\\PSNR$\uparrow$}} \\
        \midrule

        RealTime4DGS~\cite{yang2024real}
        & \DynVal{25.85} & \HighVal{\underline{24.83}}
        & \DynVal{24.37} & \HighVal{25.13}
        & \DynVal{24.82} & \HighVal{\underline{24.60}}
        & \DynVal{26.61} & \HighVal{\underline{24.61}}
        & \DynVal{25.72} & \HighVal{23.31}
        & \DynVal{24.55} & \HighVal{\underline{19.30}}
        & \DynVal{25.32} & \HighVal{\underline{23.63}} \\

        Ex4DGS~\cite{lee2024fully}
        & \DynVal{22.70} & \HighVal{18.92}
        & \DynVal{24.67} & \HighVal{24.02}
        & \DynVal{23.70} & \HighVal{18.65}
        & \DynVal{24.98} & \HighVal{21.42}
        & \DynVal{20.04} & \HighVal{17.04}
        & \DynVal{21.25} & \HighVal{16.77}
        & \DynVal{22.89} & \HighVal{19.47} \\

        4DGaussian~\cite{wu20244d}
        & \DynVal{22.68} & \HighVal{19.49}
        & \DynVal{21.60} & \HighVal{21.83}
        & \DynVal{23.36} & \HighVal{18.90}
        & \DynVal{23.56} & \HighVal{22.46}
        & \DynVal{19.57} & \HighVal{17.97}
        & \DynVal{21.83} & \HighVal{16.89}
        & \DynVal{22.10} & \HighVal{19.59} \\

        Multi4D~\cite{wang2026multi4d}
        & \DynVal{22.10} & \HighVal{20.54}
        & \DynVal{22.32} & \HighVal{21.02}
        & \DynVal{20.74} & \HighVal{17.91}
        & \DynVal{21.05} & \HighVal{15.38}
        & \DynVal{22.32} & \HighVal{18.19}
        & \DynVal{21.55} & \HighVal{16.88}
        & \DynVal{21.68} & \HighVal{18.32} \\

        FreeTimeGS~\cite{wang2025freetimegs}
        & \DynVal{\underline{26.31}} & \HighVal{24.16}
        & \DynVal{\textbf{26.47}} & \HighVal{\underline{25.91}}
        & \DynVal{\underline{26.81}} & \HighVal{24.28}
        & \DynVal{\textbf{26.92}} & \HighVal{24.30}
        & \DynVal{\underline{26.27}} & \HighVal{\underline{23.51}}
        & \DynVal{\underline{24.66}} & \HighVal{17.82}
        & \DynVal{\underline{26.24}} & \HighVal{23.33} \\

        \midrule
        \textbf{Ours}
        & \DynVal{\textbf{26.98}} & \HighVal{\textbf{25.19}}
        & \DynVal{\underline{26.26}} & \HighVal{\textbf{26.47}}
        & \DynVal{\textbf{27.52}} & \HighVal{\textbf{26.71}}
        & \DynVal{\underline{26.81}} & \HighVal{\textbf{26.61}}
        & \DynVal{\textbf{27.57}} & \HighVal{\textbf{25.87}}
        & \DynVal{\textbf{26.56}} & \HighVal{\textbf{22.93}}
        & \DynVal{\textbf{26.95}} & \HighVal{\textbf{25.63}} \\

        \midrule

        Method
        & DSSIM-1$\downarrow$ & DSSIM-2$\downarrow$
        & DSSIM-1$\downarrow$ & DSSIM-2$\downarrow$
        & DSSIM-1$\downarrow$ & DSSIM-2$\downarrow$
        & DSSIM-1$\downarrow$ & DSSIM-2$\downarrow$
        & DSSIM-1$\downarrow$ & DSSIM-2$\downarrow$
        & DSSIM-1$\downarrow$ & DSSIM-2$\downarrow$
        & DSSIM-1$\downarrow$ & DSSIM-2$\downarrow$ \\
        \midrule

        RealTime4DGS~\cite{yang2024real}
        & 0.045 & 0.023
        & 0.046 & 0.024
        & 0.043 & 0.022
        & 0.040 & 0.020
        & 0.043 & 0.022
        & 0.041 & 0.021
        & 0.043 & 0.022 \\
        
        Ex4DGS~\cite{lee2024fully}
        & 0.065 & 0.034
        & 0.058 & 0.029
        & 0.062 & 0.032
        & 0.059 & 0.030
        & 0.060 & 0.032
        & 0.099 & 0.065
        & 0.067 & 0.037 \\
        
        4DGaussian~\cite{wu20244d}
        & 0.054 & 0.028
        & 0.054 & 0.029
        & 0.052 & 0.026
        & 0.048 & 0.024
        & 0.054 & 0.028
        & 0.051 & 0.026
        & 0.052 & 0.027 \\
        
        Multi4D~\cite{wang2026multi4d}
        & 0.053 & 0.027
        & 0.051 & 0.026
        & 0.053 & 0.027
        & 0.052 & 0.027
        & 0.048 & 0.024
        & 0.051 & 0.026
        & 0.052 & 0.026 \\
        
        FreeTimeGS~\cite{wang2025freetimegs}
        & \underline{0.039} & \underline{0.019}
        & \underline{0.039} & \underline{0.020}
        & \underline{0.037} & \underline{0.018}
        & \underline{0.032} & \underline{0.015}
        & \underline{0.037} & \underline{0.018}
        & \underline{0.035} & \underline{0.017}
        & \underline{0.037} & \underline{0.018} \\
        
        \midrule
        \textbf{Ours}
        & \textbf{0.035} & \textbf{0.017}
        & \textbf{0.032} & \textbf{0.016}
        & \textbf{0.032} & \textbf{0.015}
        & \textbf{0.030} & \textbf{0.014}
        & \textbf{0.030} & \textbf{0.014}
        & \textbf{0.030} & \textbf{0.014}
        & \textbf{0.032} & \textbf{0.015} \\
        
        \bottomrule
    \end{tabular}%
    }
\end{table*}

\begin{table*}[t]
    \centering
    \footnotesize
    \caption{Scene-wise quantitative comparison on the Custom dataset.}
    \label{tab:custom_scene_quantitative}
    \setlength{\tabcolsep}{6pt}
    \renewcommand{\arraystretch}{1.12}
    \begin{tabular*}{\textwidth}{@{\extracolsep{\fill}}l*{3}{cc}@{}}
        \toprule
        \multicolumn{7}{c}{\textbf{Custom}} \\
        \midrule
        & \multicolumn{2}{c}{Baseball} & \multicolumn{2}{c}{Break} & \multicolumn{2}{c}{Avg.} \\
        \cmidrule(lr){2-3}\cmidrule(lr){4-5}\cmidrule(l){6-7}

        Method & PSNR$\uparrow$ & LPIPS$\downarrow$ & PSNR$\uparrow$ & LPIPS$\downarrow$ & PSNR$\uparrow$ & LPIPS$\downarrow$ \\
        \midrule
        RealTime4DGS~\cite{yang2024real}
        & 20.27 & 0.230
        & 20.65 & 0.268
        & 20.46 & 0.249 \\
        Ex4DGS~\cite{lee2024fully}
        & 19.61 & 0.310
        & 19.55 & 0.301
        & 19.58 & 0.306 \\
        4DGaussian~\cite{wu20244d}
        & 20.55 & 0.300
        & 20.31 & 0.397
        & 20.43 & 0.348 \\
        Multi4D~\cite{wang2026multi4d}
        & \underline{20.91} & 0.192
        & \underline{20.87} & 0.214
        & \underline{20.89} & 0.203 \\
        FreeTimeGS~\cite{wang2025freetimegs}
        & 20.79 & \underline{0.146}
        & 20.82 & \underline{0.193}
        & 20.80 & \underline{0.169} \\
        \midrule
        \textbf{Ours}
        & \textbf{20.95} & \textbf{0.141}
        & \textbf{21.05} & \textbf{0.172}
        & \textbf{21.00} & \textbf{0.157} \\
        \midrule

        Method & \DynVal{\makecell{Dynamic\\PSNR$\uparrow$}} & \HighVal{\makecell{Highly Dynamic\\PSNR$\uparrow$}} & \DynVal{\makecell{Dynamic\\PSNR$\uparrow$}} & \HighVal{\makecell{Highly Dynamic\\PSNR$\uparrow$}} & \DynVal{\makecell{Dynamic\\PSNR$\uparrow$}} & \HighVal{\makecell{Highly Dynamic\\PSNR$\uparrow$}} \\
        \midrule
        RealTime4DGS~\cite{yang2024real}
        & \DynVal{24.18} & \HighVal{19.57}
        & \DynVal{24.02} & \HighVal{20.27}
        & \DynVal{24.10} & \HighVal{19.92} \\
        Ex4DGS~\cite{lee2024fully}
        & \DynVal{21.60} & \HighVal{18.59}
        & \DynVal{22.52} & \HighVal{16.31}
        & \DynVal{22.06} & \HighVal{17.45} \\
        4DGaussian~\cite{wu20244d}
        & \DynVal{20.32} & \HighVal{16.56}
        & \DynVal{22.70} & \HighVal{13.70}
        & \DynVal{21.51} & \HighVal{15.13} \\
        Multi4D~\cite{wang2026multi4d}
        & \DynVal{25.25} & \HighVal{21.72}
        & \DynVal{21.91} & \HighVal{17.58}
        & \DynVal{23.58} & \HighVal{19.65} \\
        FreeTimeGS~\cite{wang2025freetimegs}
        & \DynVal{\underline{27.53}} & \HighVal{\underline{25.62}}
        & \DynVal{\underline{27.67}} & \HighVal{\underline{22.52}}
        & \DynVal{\underline{27.60}} & \HighVal{\underline{24.07}} \\
        \midrule
        \textbf{Ours}
        & \DynVal{\textbf{27.91}} & \HighVal{\textbf{27.04}}
        & \DynVal{\textbf{27.75}} & \HighVal{\textbf{26.70}}
        & \DynVal{\textbf{27.83}} & \HighVal{\textbf{26.87}} \\
        \midrule

        Method & DSSIM-1$\downarrow$ & DSSIM-2$\downarrow$ & DSSIM-1$\downarrow$ & DSSIM-2$\downarrow$ & DSSIM-1$\downarrow$ & DSSIM-2$\downarrow$ \\
        \midrule
        RealTime4DGS~\cite{yang2024real}
        & 0.152 & 0.110 & 0.150 & \underline{0.100} & 0.151 & 0.105 \\
        Ex4DGS~\cite{lee2024fully}
        & 0.169 & 0.125 & 0.177 & 0.129 & 0.173 & 0.127 \\
        4DGaussian~\cite{wu20244d}
        & 0.159 & 0.112 & 0.197 & 0.136 & 0.178 & 0.124 \\
        Multi4D~\cite{wang2026multi4d}
        & 0.142 & 0.106
        & 0.159 & 0.116
        & 0.151 & 0.111 \\
        FreeTimeGS~\cite{wang2025freetimegs}
        & \underline{0.125} & \underline{0.094}
        & \underline{0.138} & 0.101
        & \underline{0.132} & \underline{0.098} \\
        \midrule
        \textbf{Ours}
        & \textbf{0.122} & \textbf{0.092}
        & \textbf{0.131} & \textbf{0.097}
        & \textbf{0.127} & \textbf{0.095} \\

        \bottomrule
    \end{tabular*}
\end{table*}

\begin{figure*}[t]
    \centering
    \includegraphics[width=\linewidth]{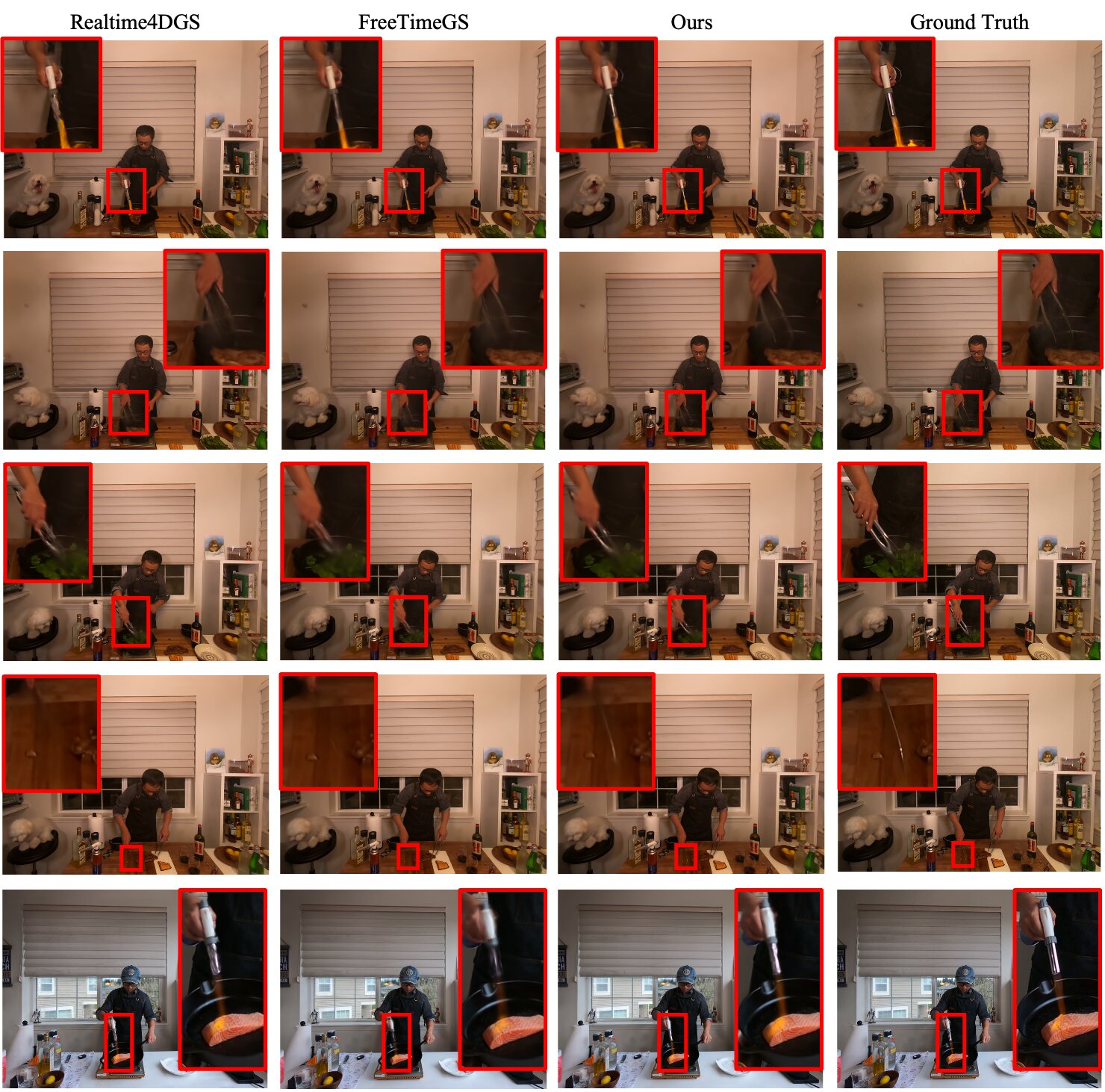}
    \caption{
        \textbf{Additional qualitative comparisons on N3DV.}
        Our method better preserves fine details in dynamic regions, including the thin knife in the fourth row.
    }
    \label{fig:n3dv_qualitative}
\end{figure*}

\end{document}